\documentclass[sn-mathphys-num]{sn-jnl}
\usepackage{graphicx}
\usepackage{amssymb}
\usepackage{booktabs}
\usepackage{siunitx}
\usepackage{amsmath}
\usepackage{multirow}
\usepackage{hyperref}
\usepackage{xcolor}
\usepackage{float}
\usepackage{placeins} 
\usepackage{tabularx}
\usepackage{array}
\usepackage[table]{xcolor}

\usepackage{rotating}

\title[Title]{PULSE-ICU: A Pretrained Unified Long-Sequence Encoder for Multi-task Prediction in Intensive Care Units}

\author[1]{\fnm{Sejeong} \sur{Jang}}
\author*[2]{\fnm{Joo Heung} \sur{Yoon} \email{yoonjh@upmc.edu}}
\author*[1]{\fnm{Hyo Kyung} \sur{Lee} \email{hyokyunglee@korea.ac.kr}}

\affil[1]{\orgdiv{School of Industrial and Management Engineering}, \orgname{Korea University}, \orgaddress{\city{Seoul}, \country{Republic of Korea}}}

\affil[2]{\orgdiv{Division of Pulmonary, Allergy, Critical Care, and Sleep Medicine, Department of Medicine}, \orgname{University of Pittsburgh}, \orgaddress{\city{Pittsburgh, PA},\country{USA}}}

\abstract{Intensive care unit (ICU) data are highly irregular, heterogeneous, and temporally fragmented, posing challenges for generalizable clinical prediction. We present PULSE-ICU, a self-supervised foundation model that learns event-level ICU representations from large-scale EHR sequences without resampling or manual feature engineering. A unified embedding module encodes event identity, continuous values, units, and temporal attributes, while a Longformer-based encoder enables efficient modeling of long trajectories.
PULSE-ICU was fine-tuned across 18 prediction tasks, including mortality, intervention forecasting, and phenotype identification, achieving strong performance across task types. External validation on eICU, HiRID, and P12 showed substantial improvements with minimal fine-tuning, demonstrating robustness to domain shift and variable constraints.
These findings suggest that foundation-style modeling can improve data efficiency and adaptability, providing a scalable framework for ICU decision support across diverse clinical environments.}


\begin{document}
\maketitle
\section{Introduction}

\begin{figure*}[ht]
    \centering
    \includegraphics[width=0.9\textwidth]{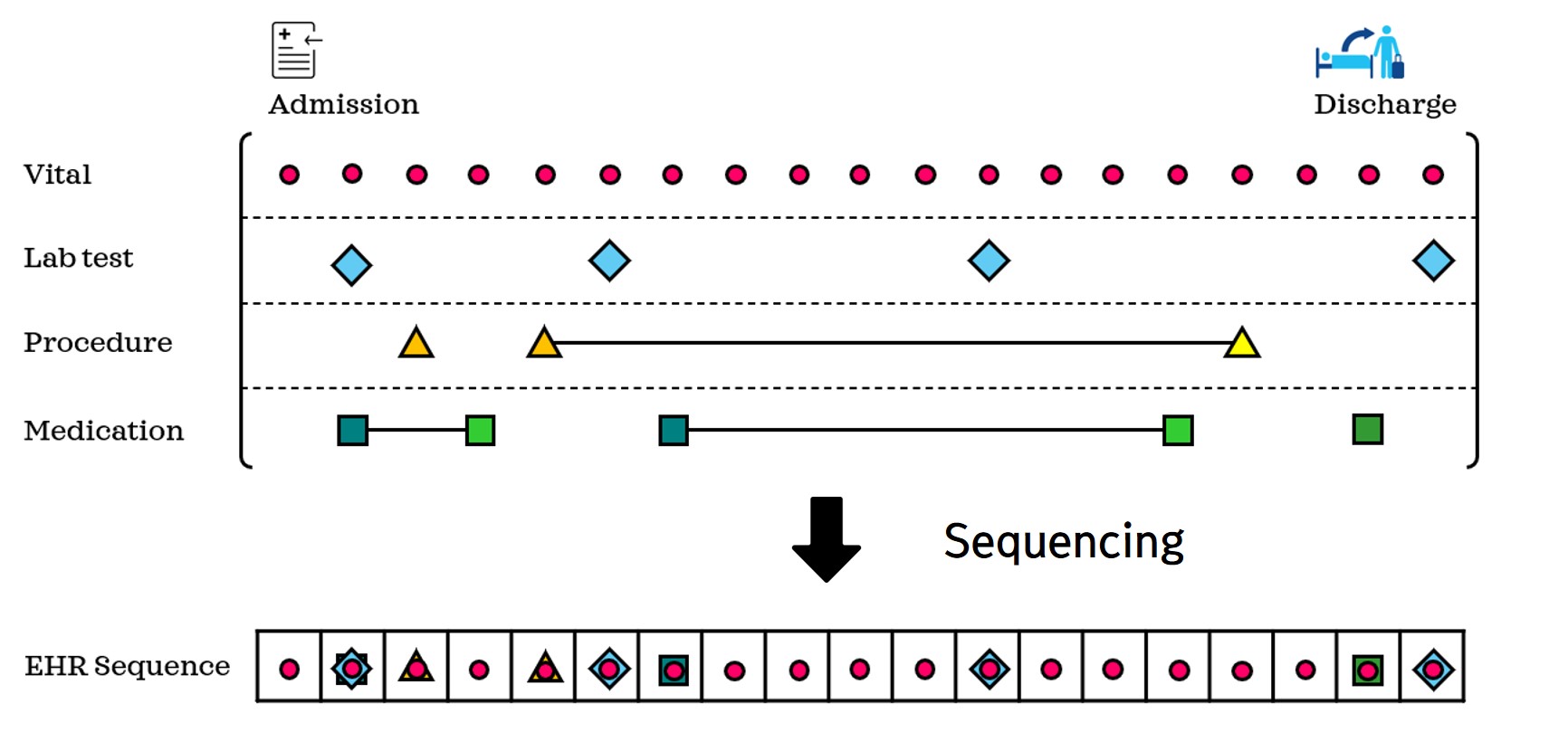}
    \caption{Illustration of irregular, heterogeneous ICU event sequences in EHR. Each row corresponds to a clinical variable (vital sign, laboratory test, procedure, medication), and each dot denotes a recorded event at its native timestamp. 
}
    \label{figure : ehr figure}
\end{figure*}

Patients admitted to the intensive care unit (ICU) generate large volumes of heterogeneous, fine-grained clinical data, including vital signs, laboratory measurements, administered medications, and procedural interventions. These events are recorded at highly irregular intervals and with variable measurement resolutions. As illustrated in Figure~\ref{figure : ehr figure}, some physiological variables are captured at high frequency, whereas others occur only intermittently, producing complex temporal structures that challenge conventional modeling pipelines. Standard preprocessing approaches—such as coarse resampling, summary aggregation, or feature engineering based on hand-crafted heuristics—often distort the underlying physiological dynamics and diminish the model’s ability to detect subtle but clinically meaningful trends in patient trajectories.

\FloatBarrier

Early studies on structured electronic health record (EHR) data predominantly adopted classical machine-learning models, such as logistic regression and random forests~\cite{breiman2001random}, developed around manually engineered predictors. The emergence of deep learning introduced recurrent architectures, including LSTMs~\cite{hochreiter1997long} and GRUs~\cite{chung2014empirical}, which enabled sequential modeling of ICU time-series data. RETAIN~\cite{choi2016retain} extended this paradigm by introducing a reverse-time attention mechanism that enhanced interpretability while maintaining predictive performance.

Self-supervised learning further transformed EHR modeling, progressing along two main directions~\cite{vaswani2017attention}. The first focuses on unstructured clinical narratives, where models such as ClinicalBERT~\cite{huang2019clinicalbert}, Clinical-BigBird, and Clinical-Longformer~\cite{li2022clinical} leverage large text corpora to learn contextualized representations that generalize across diverse downstream clinical tasks. The second direction targets structured clinical event sequences. BEHRT~\cite{li2020behrt} introduced a visit-level Transformer to model longitudinal diagnosis trajectories, later extended by Hi-BEHRT~\cite{li2022hi} to jointly capture intra-visit and inter-visit dependencies. ExBEHRT~\cite{rupp2023exbehrt} incorporated additional modalities such as medications and laboratory results, while HEART~\cite{huang2024heart} enhanced intra-visit reasoning through relation-aware attention. Collectively, these models demonstrate the value of Transformer architectures in representing long-term clinical histories, but they typically operate at the visit or encounter level—an abstraction that does not capture the fine temporal granularity characteristic of ICU data.

As summarized in Table \ref{tab:related_foundation_models}, several recent efforts have explored self-supervised pretraining directly on ICU-level structured event streams. GenHPF~\cite{hur2023genhpf} introduced a hierarchical representation of ICU events using text-based embeddings combined with SimCLR and masked-language objectives. DuETT~\cite{labach2023duett} encoded each event as a triplet (event type, value, timestamp) and proposed an event–time dual-Transformer optimized through masked event prediction and value reconstruction. ICU-BERT~\cite{santos2025improving} extended this idea with a quadruplet embedding incorporating feature identity, continuous value, unit, and temporal attributes, trained with a unified masked language–value modeling objective. Other recent studies similarly combine event-type, temporal, and value embeddings with masked language modeling (MLM) or contrastive objectives to build self-supervised EHR models~\cite{xu2023transehr, king2023multimodal, king2024efficient, jin2025novel, jagd2025towards}.

\begin{sidewaystable}[!htbp]

\centering
\caption{Representative ICU-focused foundation models for structured EHR data. For each study, we summarize the primary datasets, event embedding strategy, self-supervised pretraining objectives, backbone architecture, and major downstream clinical applications.}

\renewcommand{\arraystretch}{1.25}  
\setlength{\tabcolsep}{4pt}         

{\footnotesize 
\begin{tabularx}{\textheight}{l X X X X X}
\toprule
\textbf{Paper (year)} & \textbf{Dataset} &
\textbf{Embedding Methods} & \textbf{Pretraining Task} &
\textbf{Architecture} & \textbf{Downstream Task} \\
\midrule

GenHPF~\cite{hur2023genhpf} (2022) &
MIMIC-III, MIMIC-IV, eICU &
Text-based embedding &
SimCLR &
Hierarchical Transformer Encoder &
Mortality, LOS, Readmission, Final Acuity, Imminent Discharge, Diagnosis, Lab Values, etc. (Multitask) \\

TransEHR~\cite{xu2023transehr} (2023) &
MIMIC-III, Physionet-12, AmsterdamUMCdb &
Triplet Embedding (Categorical, Continuous, Temporal) &
MLM &
Transformer Encoder &
In-hospital mortality, LOS, Phenotyping \\

King et al.~\cite{king2023multimodal} (2023) &
MIMIC-III &
Joint embedding (measurements, notes) &
MLM + Alignment &
BERT-based Encoder &
In-hospital Mortality, Phenotype \\

DuETT~\cite{labach2023duett} (2023) &
MIMIC-IV, PhysioNet 2012 &
Triplet Embedding (Event-type, Time, Value) &
Masked Event and Value Prediction &
Event-Time Dual Transformer &
In-hospital Mortality, Phenotype \\

King et al.~\cite{king2024efficient} (2024) &
MIMIC-III, eICU &
Triplet Embedding (Event-type, Time, Value) &
MLM + Contrastive &
Transformer Encoder &
In-hospital Mortality, Phenotype, Imputation \\

ICU-BERT~\cite{santos2025improving} (2025) &
MIMIC-IV, YAIB &
Quadruplet Multi-token Embedding &
Masked Language-Value Modeling &
BERT &
In-hospital Mortality, Phenotype, ICU Mortality, AKI Onset, Kidney Function \\

Jin et al.~\cite{jin2025novel} (2025) &
MIMIC-IV &
MAE Encoder-Decoder &
MAE &
Teacher-Student Multitask &
Mortality Classification, Severity Score Regression \\

Jagd et al.~\cite{jagd2025towards} (2025) &
MIMIC-III, MIMIC-IV, eICU &
Triplet Embedding (Feature Identity, Continuous, Temporal) &
Forecasting &
Bi-Axial Transformer &
Mortality \\
\bottomrule

\end{tabularx}
} 
\label{tab:related_foundation_models}
\end{sidewaystable}

Beyond these Transformer-based efforts, alternative pretraining paradigms have emerged. ETHOS~\cite{renc2024zero} tokenizes longitudinal EHR timelines into structured event sequences and demonstrates zero-shot capabilities. CEHR-GPT~\cite{pang2024cehr} unifies multitask learning, zero-shot inference, and synthetic data generation, while EHR-Mamba~\cite{fallahpour2024ehrmamba} employs state-space architectures to improve efficiency for very long sequences. Taken together, these studies highlight the rapid evolution of foundation-model approaches for structured clinical data.

Despite this progress, three key challenges remain. First, most Transformer-based models struggle to process raw ICU event streams due to the quadratic cost of self-attention, necessitating truncation or coarse regularization of long sequences. Second, continuous physiological values are typically embedded using straightforward normalization, binning, or linear projection techniques, which inadequately capture the complex and diverse numerical characteristics inherent in ICU measurements. Third, prior pretraining studies commonly evaluate performance on a small set of downstream tasks in isolation, limiting their applicability as general-purpose clinical reasoning models.

To address these limitations, we propose PULSE-ICU (\textit{Pretrained Unified Long-Sequence Encoder for multi-task prediction in Intensive Care Units}), a foundation-style ICU modeling framework that directly encodes more than 900 types of clinical events at their native temporal resolution without resampling. PULSE-ICU employs a Longformer-based sparse-attention encoder to efficiently model long and irregular event sequences while preserving global contextual information. In addition, it introduces an attribute-specific multi-embedding architecture that separately represents event identity, continuous values, measurement units, and temporal properties, enabling faithful modeling of heterogeneous ICU data distributions. Finally, we perform large-scale multi-task fine-tuning across 18 prediction tasks, including mortality, length of stay, readmission, phenotype classification, organ dysfunction scores, and intervention-related outcomes. PULSE-ICU provides a unified, scalable, and clinically relevant framework for ICU event modeling.

\section{Results}

\subsection{Cohort characteristics and experimental setup}

We utilized the MIMIC-IV database for both self-supervised pretraining and multi-task fine-tuning, and employed YAIB-aligned external datasets (HiRID, eICU) as well as PhysioNet 2012 (P12) for domain adaptation and external validation. Table~\ref{tab:dataset_statistics_comparison_column} summarizes the key demographic and temporal characteristics of all cohorts analyzed in this study.

The pretraining cohort (MIMIC-PT) comprised 57,726 ICU stays and 951 unique clinical variables, capturing high-dimensional event streams from heterogeneous data sources within the ICU. The fine-tuning cohort (MIMIC-FT) adopted the same variable schema but consisted of shorter 24-hour observation windows tailored for supervised downstream learning.
To assess the practical adaptability of the model, we additionally constructed a feature-limited cohort (MIMIC-Limited) containing 72 core clinical variables. These variables included YAIB-defined vital signs and laboratory tests, as well as Glasgow Coma Scale scores, vasopressor administration, and mechanical ventilation indicators. This setting enabled evaluation of model performance under restricted-variable conditions that more closely reflect routine clinical environments, where only a subset of measurements is consistently available.

For external validation, we adopted the YAIB variable schema to harmonize feature definitions across HiRID and eICU. The HiRID cohort consisted of 31,517 ICU stays with 38 aligned variables and exhibited long, densely sampled temporal sequences (mean 3,471 events and 24 unique variables per stay). The eICU cohort included 127,549 ICU stays from multiple hospitals, characterized by shorter and sparser sequences (mean 215 events and 25 unique variables per stay). The P12 cohort contained 11,667 ICU stays with 34 variables collected over 48-hour observation windows, with an average of 397 events and 25 unique variables per sequence.

Label distributions for each fine-tuning task are provided in Tables~\ref{tab:mimic_label_distribution}, \ref{tab:eicu_label_distribution}, \ref{tab:hirid_label_distribution}, and~\ref{tab:p12_label_distribution} in the Appendix.

\begin{table*}[ht]
\centering
\caption{Cohort characteristics across pretraining and fine-tuning datasets. MIMIC-PT denotes the full-length ICU cohort used for self-supervised pretraining, whereas MIMIC-FT and MIMIC-Limited correspond to fine-tuning cohorts with full and YAIB-core variable sets, respectively. HiRID-YAIB and eICU-YAIB are external validation cohorts aligned to the YAIB schema.}
\renewcommand{\arraystretch}{1.2}
\resizebox{0.95\textwidth}{!}{
\begin{tabular}{@{}lcccccc@{}}
\toprule
\textbf{Characteristic} & \textbf{MIMIC-PT} & \textbf{MIMIC-FT} & \textbf{MIMIC-Limited} & \textbf{HiRID-YAIB} & \textbf{eICU-YAIB}  & \textbf{P12} \\ 
\midrule
Mean Age &   65   &  65   &    65     &     63    &    64   &   65    \\
Male (n, \%)  &   32,636 (57\%)   &   32,618 (57\%)  &    32,613 (57\%)     &    20,340 (65\%)     &   58,296 (46\%)    &   6,602 (56\%)   \\
No. ICU Stays & 57,726 (69,247) & 57,693 & 57,688 & 31,517  & 127,549 & 11,667 \\
No. Unique Events & 951 & 951 & 72 & 38 & 47 & 34 \\
Avg. Sequence Length & 1,795 & 604 & 256 & 3,471  & 215 & 397 \\
Avg. Unique Events per Sequence & 154 & 134 & 35 & 24 &  25 &  25 \\
\bottomrule
\end{tabular}
}
\label{tab:dataset_statistics_comparison_column}
\end{table*}

\subsection{Pretraining and multi-task fine-tuning performance}

\paragraph{Pretraining performance and masking strategy}

We first assessed the effectiveness of the self-supervised pretraining strategy employed in PULSE-ICU by evaluating masked event prediction (MEP) and value prediction (VP) across six masking configurations. Table~\ref{tab:Pretrain Results} summarizes the precision for each event type—Medication, Chart, and Procedure—and the corresponding value prediction loss.

Chart events, which constitute the majority of tokens in the pretraining corpus, achieved the highest prediction precision across all masking ratios. Medication and Procedure events, despite being considerably less frequent, were also predicted with moderate accuracy, indicating that the model successfully captured cross-event dependencies and learned a coherent latent structure of heterogeneous ICU event streams. Procedure events, in particular, accounted for less than 1\% of all tokens yet remained consistently predictable, suggesting that the embedding architecture and sparse-attention mechanism effectively modeled rare event types.

To investigate how the masking strategy affected downstream task performance, we compared fine-tuning results across 18 clinical prediction tasks using the same six masking configurations. As shown in Table~\ref{tab:masking_ratio_results}, the 30/15/30/5 and 20/15/20/15 configurations achieved the highest overall AUROC and AUPRC when averaged across all task categories. However, all masking strategies demonstrated pronounced variability in the SOFA–Coagulation task, with large fluctuations across runs. Excluding this unstable task from aggregation, the 30/30/30/5 configuration consistently achieved the best overall performance across binary, multi-class, and multi-label prediction tasks. Thus, we selected this masking configuration as the final pretraining strategy for all subsequent experiments.

\begin{table*}[ht]
    \centering
    \caption{Self-supervised pretraining performance under different masking configurations. 
Each masking ratio (M/C/P/V) specifies the proportion of masked Medication, Chart, Procedure, and Value events, respectively, used in the masked event prediction (MEP) and value prediction (VP) objectives.}
    \vspace{0.5em}  
    \resizebox{\textwidth}{!}{  
    \begin{tabular}{@{}lccccc@{}}
        \toprule
        \textbf{Masking Ratio} & \textbf{Medication Precision} & \textbf{Chart Precision} & \textbf{Procedure Precision} & \textbf{All Precision} & \textbf{Value Prediction Loss}\\ \midrule
        30/30/30/00 & 0.537 & 0.834 & 0.471 & 0.801 & -\\
        30/30/30/05 & 0.529 & 0.812 & 0.473 & 0.781 & 28.36\\
        30/30/30/10 & 0.535 & 0.816 & 0.473 & 0.785 & 29.74\\
        30/30/30/15 & 0.533 & 0.823 & 0.474 & 0.792 & 29.67\\ 
        30/15/30/05 & 0.559 & 0.829 & 0.486 & 0.776 & 27.99\\
        20/15/20/15 & 0.588 & 0.843 & 0.597 & 0.808 & 29.24\\
        \bottomrule
    \end{tabular}
    }
    \label{tab:Pretrain Results}
\end{table*}

\begin{table*}[ht]
    \centering
    \caption{Downstream multi-task performance of PULSE-ICU under different pretraining masking configurations. Each masking ratio (M/C/P/V) indicates the masking rates for Medication, Chart, Procedure, and Value events during pretraining. Mean AUROC and AUPRC are reported for all 18 tasks combined (All) and separately for binary, multi-class (SOFA), and multi-label (phenotype) task groups.}
    \smallskip
    \renewcommand{\arraystretch}{1.15}
    \resizebox{\textwidth}{!}{
    \begin{tabular}{@{\hspace{1em}}lcccccccccccc@{}}
        \toprule
        \multirow{2}{*}{\textbf{Task Group}} &
        \multicolumn{2}{c}{\textbf{30/30/30/0}} & 
        \multicolumn{2}{c}{\textbf{30/30/30/5}} & 
        \multicolumn{2}{c}{\textbf{30/30/30/10}} &
        \multicolumn{2}{c}{\textbf{30/30/30/15}} &
        \multicolumn{2}{c}{\textbf{30/15/30/5}} & 
        \multicolumn{2}{c}{\textbf{20/15/20/15}} \\
        \cmidrule(lr){2-13}
        & AUROC & AUPRC & AUROC & AUPRC & AUROC & AUPRC & AUROC & AUPRC & AUROC & AUPRC & AUROC & AUPRC \\ 
        \midrule
        \textbf{All}         & 0.860 & 0.484 & \textbf{0.865} & \textbf{0.496} & 0.859 & 0.488 & 0.866 & 0.491 & 0.865 & 0.499 & 0.866 & 0.497 \\
        \textbf{Binary}      & 0.855 & 0.451 & \textbf{0.863} & \textbf{0.462} & 0.856 & 0.451 & 0.859 & 0.449 & 0.858 & 0.460 & 0.858 & 0.456 \\
        \textbf{Multi-Class} & 0.886 & 0.550 & \textbf{0.882} & \textbf{0.559} & 0.878 & 0.557 & 0.895 & 0.571 & 0.895 & 0.576 & 0.898 & 0.576 \\
        \textbf{Multi-Label} & 0.750 & 0.453 & \textbf{0.776} & \textbf{0.493} & 0.775 & 0.489 & 0.767 & 0.480 & 0.767 & 0.477 & 0.769 & 0.481 \\
        \bottomrule
    \end{tabular}
    }
    \label{tab:masking_ratio_results}
\end{table*}

\paragraph{Multi-task fine-tuning performance on MIMIC-IV}
\label{sec:multi-task-fine-tuning-results}

Using the pretrained PULSE-ICU model, we conducted multi-task fine-tuning across 18 clinical prediction tasks, encompassing binary, multi-class, and multi-label outcomes. For each task, AUROC and AUPRC were computed, and for multi-class and multi-label settings, both macro- and micro-averaged metrics were reported to provide a comprehensive assessment of model behavior. All values represent the mean~$\pm$~standard deviation from 5-fold cross-validation. The full results are summarized in Table~\ref{tab:5fold_results}.

Overall, the fine-tuned PULSE-ICU model demonstrated strong and stable predictive performance across heterogeneous task types. In binary tasks, PULSE-ICU achieved an AUROC of 0.887 for in-hospital mortality and 0.932 for ICU mortality, with corresponding AUPRCs of 0.550 and 0.618, illustrating its effectiveness in early mortality risk estimation. Tasks with substantial class imbalance, such as 12-hour ventilation prediction (positive prevalence $<$1\%), yielded lower AUPRC values, which is expected given the rarity of positive cases.

Performance was also consistent across SOFA-based multi-class tasks. Liver and renal function predictions achieved AUROC values of 0.931 and 0.926, respectively, indicating that the model effectively captured physiological signatures associated with organ dysfunction. Notably, respiratory SOFA achieved a micro-AUROC of 0.987, reflecting PULSE-ICU's ability to model dense and high-frequency respiratory events.

In the multi-label phenotype prediction task, PULSE-ICU attained a Macro-AUROC of 0.785 and a Micro-AUROC of 0.832 using only the first 24 hours of ICU data. These results demonstrate that the learned representations support concurrent identification of diverse disease phenotypes from limited early-stage information.

\begin{figure*}[ht]
    \centering
    \includegraphics[width=0.9\textwidth]{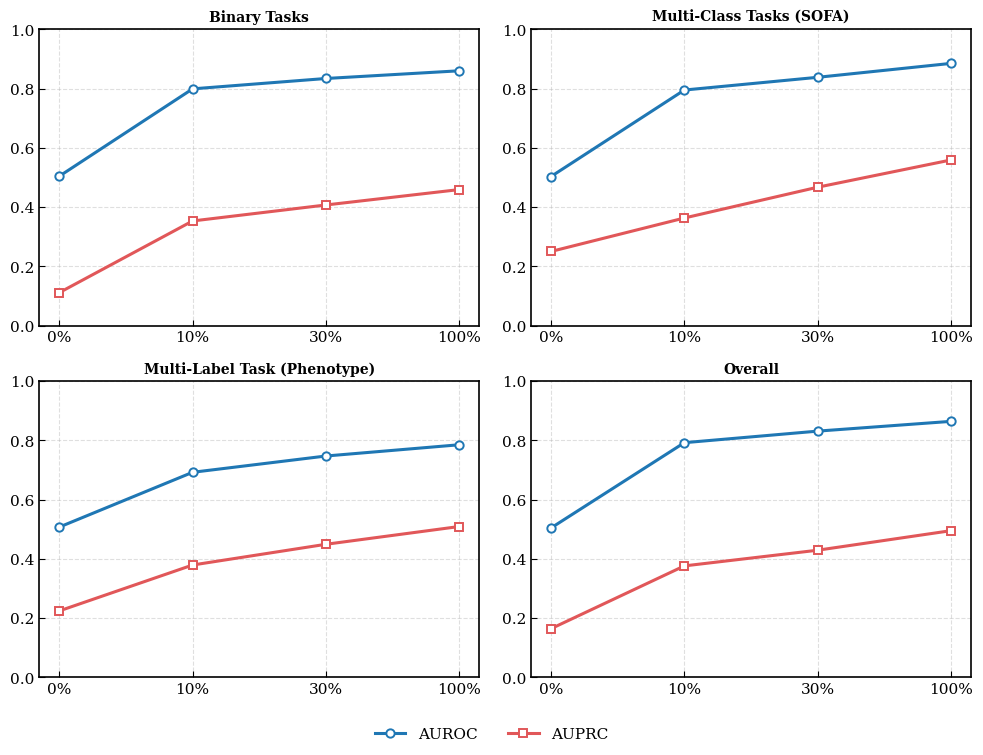}
    \caption{Effect of fine-tuning data size on the downstream performance of PULSE-ICU in MIMIC-IV. Mean AUROC and AUPRC are plotted for binary classification, multi-class (SOFA; macro-averaged), multi-label (phenotype; macro-averaged), and the overall average across all tasks as the proportion of labeled fine-tuning data increases (0\%, 10\%, 30\%, and 100\%).}
    \label{figure : Main Finetuning Ratio}
\end{figure*}

Figure~\ref{figure : Main Finetuning Ratio} illustrates performance trends as a function of available fine-tuning data (0\%, 10\%, 30\%, and 100\%). Across binary, multi-class, and multi-label tasks, performance improved monotonically with increasing amounts of labeled data. Even small fractions of fine-tuning data (e.g., 10–30\%) yielded substantial gains, demonstrating the data efficiency of the pretrained PULSE-ICU representations. Detailed per-task performance curves are provided in Appendix Figure~\ref{figure : Main Finetuning Ratio All Tasks}.

\begin{table*}[ht]
\centering
\caption{Five-fold cross-validation performance of the fine-tuned PULSE-ICU model on MIMIC-IV across all downstream clinical prediction tasks using a 24-hour observation window. All values are expressed as mean~$\pm$~standard deviation across folds.}
\renewcommand{\arraystretch}{1.15}
\resizebox{\textwidth}{!}{
\begin{tabular}{@{}p{4.2cm}cc|p{4.2cm}cccc@{}}
\toprule
\multicolumn{3}{c|}{\textbf{Binary Classification Task}} &
\multicolumn{5}{c}{\textbf{Multi-Class Classification Task (SOFA, 4 classes)}} \\
\midrule
\textbf{Task} & \textbf{AUROC} & \textbf{AUPRC} &
\textbf{Task} & \textbf{Macro-AUROC} & \textbf{Macro-AUPRC} & \textbf{Micro-AUROC} & \textbf{Micro-AUPRC} \\
\midrule
Mortality 30days & 0.864 ± 0.002 & 0.589 ± 0.003 & Central nervous system & 0.867 ± 0.001 & 0.605 ± 0.002 & 0.924 ± 0.000 & 0.827 ± 0.001 \\
Mortality in Hospital & 0.887 ± 0.001 & 0.550 ± 0.005 & Cardiovascular system & 0.871 ± 0.005 & 0.675 ± 0.016 & 0.910 ± 0.003 & 0.738 ± 0.010 \\
Mortality in ICU & 0.932 ± 0.002 & 0.618 ± 0.004 & Respiratory system & 0.880 ± 0.005 & 0.429 ± 0.002 & 0.987 ± 0.000 & 0.970 ± 0.000 \\
Mortality 48hr & 0.935 ± 0.004 & 0.293 ± 0.011 & Coagulation & 0.834 ± 0.037 & 0.485 ± 0.055 & 0.939 ± 0.010 & 0.857 ± 0.021 \\
ICU LOS 3 days & 0.837 ± 0.002 & 0.793 ± 0.004 & Liver & 0.931 ± 0.004 & 0.535 ± 0.025 & 0.993 ± 0.001 & 0.982 ± 0.001 \\
ICU LOS 7 days & 0.866 ± 0.001 & 0.563 ± 0.006 & Renal function & 0.926 ± 0.003 & 0.625 ± 0.007 & 0.967 ± 0.001 & 0.920 ± 0.001 \\
Readmission 30days & 0.675 ± 0.007 & 0.151 ± 0.006 &  &  &  &  &  \\
Transfusion 12hr & 0.822 ± 0.005 & 0.185 ± 0.005 &  &  &  &  &  \\
Vasopressor 12hr & 0.943 ± 0.002 & 0.773 ± 0.005 & \multicolumn{5}{c}{\textbf{Multi-Label Classification Task (25 classes)}} \\
\cmidrule(lr){4-8}
Ventilation 12hr & 0.752 ± 0.010 & 0.042 ± 0.006 & \textbf{Task} & \textbf{Macro-AUROC} & \textbf{Macro-AUPRC} & \textbf{Micro-AUROC} & \textbf{Micro-AUPRC} \\
\cmidrule(lr){4-8}
Shock 8hr & 0.945 ± 0.002 & 0.488 ± 0.023 & Phenotype & 0.785 ± 0.002 & 0.509 ± 0.003 & 0.832 ± 0.002 & 0.605 ± 0.003 \\
\bottomrule
\end{tabular}
}
\label{tab:5fold_results}
\end{table*}

\FloatBarrier

\subsection{Inference under limited-variable conditions}

Although the primary model (PULSE-ICU) was fine-tuned using the full set of clinical variables available in MIMIC-IV, real-world clinical environments often provide only a limited subset of measurements due to differences in EHR systems, documentation practices, and institutional workflows. To evaluate the practical utility of the model under such constraints, we examined its performance when inference and fine-tuning were restricted to the 72 core variables.

Following the procedure outlined in Section~\ref{sec:limited_ver}, we constructed input sequences using only these core variables and assessed the performance of the original MIMIC-finetuned PULSE-ICU model without retraining (zero-shot), as well as with varying degrees of additional fine-tuning (1\%, 5\%, 10\%, 30\%, 50\%, and 100\% of available labeled data). Zero-shot evaluations were conducted using five independently trained folds of the pretrained backbone, whereas fine-tuning experiments were repeated three times with different random seeds, with mean~$\pm$~standard deviation reported for all settings.

\begin{figure*}[ht]
    \centering
    \includegraphics[width=0.9\textwidth]{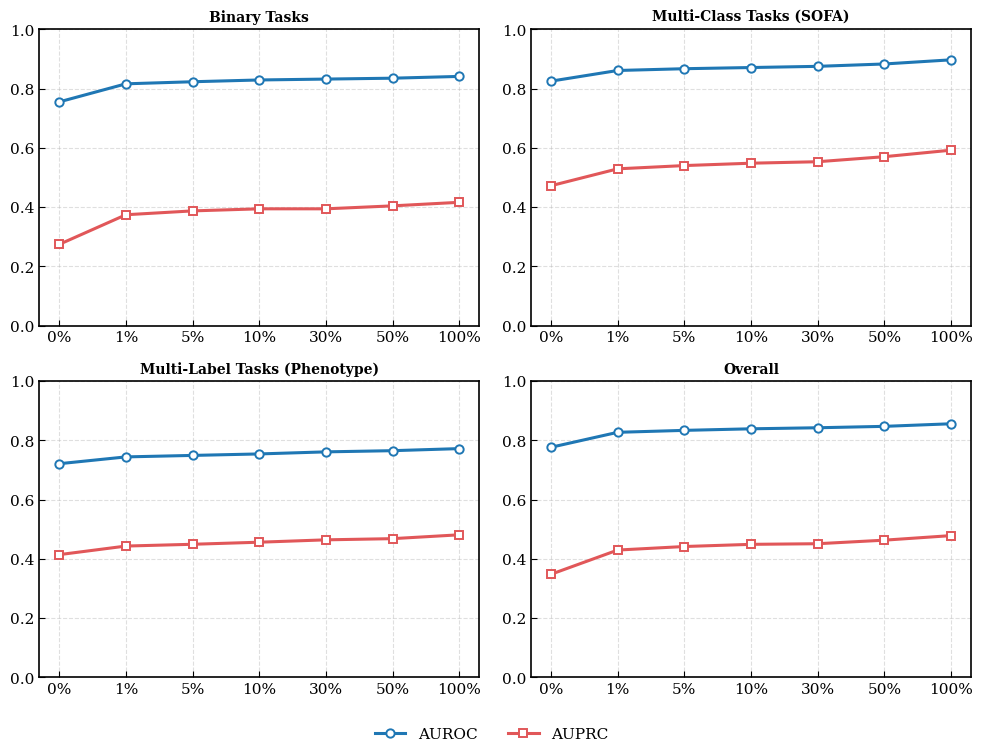}
    \caption{Performance of the fine-tuned PULSE-ICU model under feature-limited conditions (MIMIC-Limited). Using only the 72 variables, AUROC (solid blue line) and AUPRC (dashed red line) are reported for binary, multi-class (SOFA), multi-label (phenotype), and overall task groups as the proportion of labeled fine-tuning data increases from 0\% (zero-shot) to 100\%.}

    \label{figure : MIMIC limited ver}
\end{figure*}

Figure~\ref{figure : MIMIC limited ver} illustrates AUROC and AUPRC trends across binary, multi-class (SOFA), and multi-label (phenotype) prediction tasks. Overall, the model preserved substantial performance in the zero-shot setting despite the reduced feature set, demonstrating strong inherent generalization. Notably, with as little as 1\% of labeled data for fine-tuning, the model recovered a significant proportion of its full-data performance across all task groups. Performance continued to improve with additional labeled data, approaching near-original levels when using the complete fine-tuning set. Detailed per-task outcomes are provided in Appendix Figure~\ref{figure : MIMIC YAIB ver All Tasks}.

These findings highlight the robustness of the learned representations and their adaptability to variable availability constraints. Even when restricted to a condensed subset of core clinical variables within the same MIMIC-IV domain, PULSE-ICU maintained stable performance across task categories, indicating that reliable prediction is feasible under realistic, feature-limited clinical conditions.

\FloatBarrier

\subsection{External validation and domain adaptation}

\begin{figure*}[ht]
    \centering
    \includegraphics[width=0.9\textwidth]{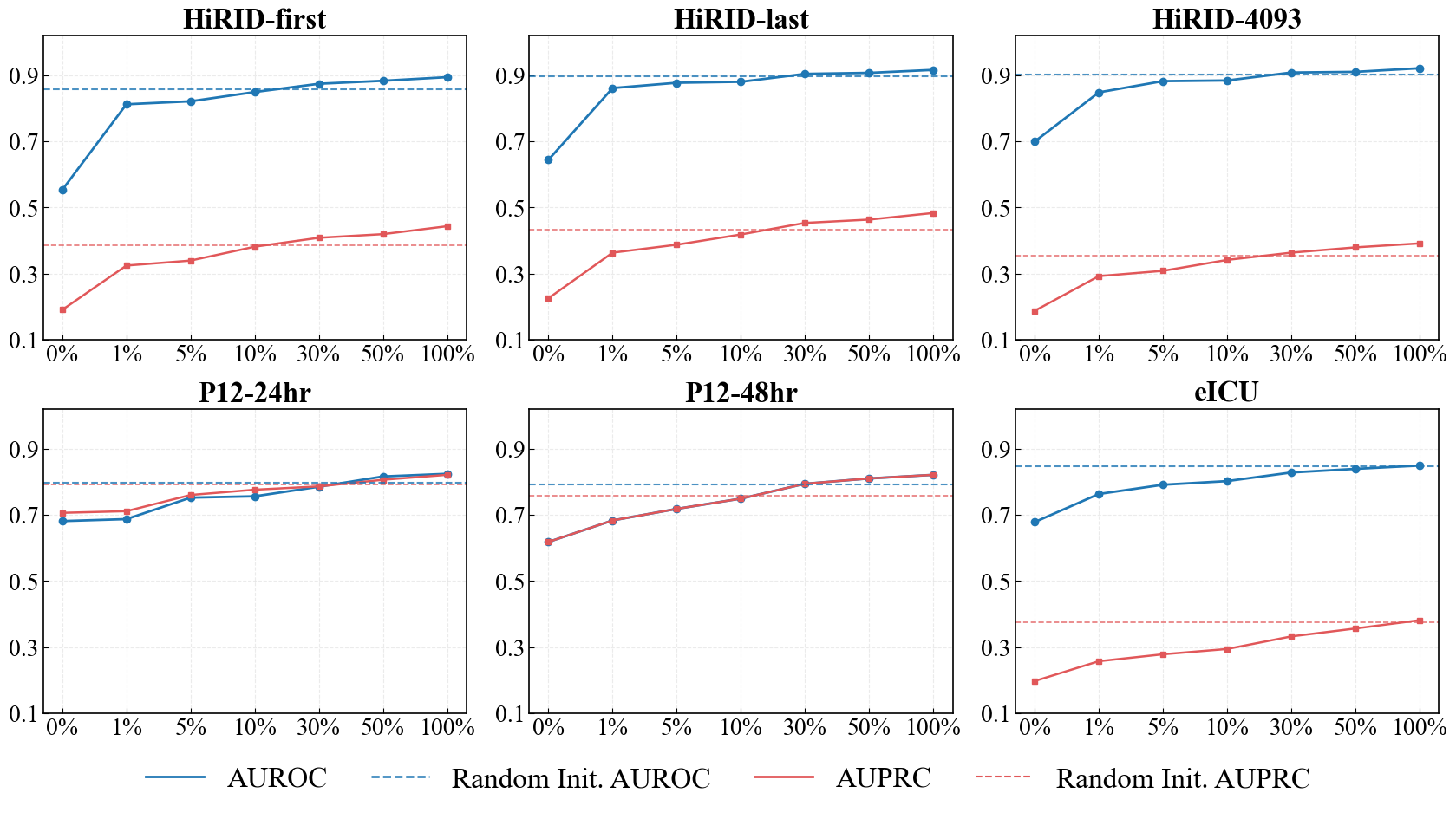}
    \caption{External validation and domain adaptation performance of the fine-tuned PULSE-ICU model across six target datasets: HiRID-first, HiRID-last, HiRID-4093, P12-24hr, P12-48hr, and eICU. For each dataset, AUROC (blue) and AUPRC (red) are plotted as functions of the proportion of target-domain data used for fine-tuning (0\%–100\%). Dashed horizontal lines denote the performance of models trained from random initialization using 100\% of the target-domain data.}
    \label{figure : external validation}
\end{figure*}

We next evaluated the transferability of the pretrained and fine-tuned PULSE-ICU model to three external ICU datasets—HiRID, eICU, and PhysioNet 2012 (P12)—representing substantial variation in clinical practice, sampling frequency, and institutional characteristics. Figure~\ref{figure : external validation} summarizes performance trends for zero-shot inference and fine-tuning with progressively larger fractions of target-domain data.

In the zero-shot setting, PULSE-ICU achieved an average AUROC of approximately 0.65 across external datasets, indicating that representations learned from MIMIC-IV retained moderate discriminative capacity despite substantial domain shift. Fine-tuning with even small amounts of labeled target-domain data led to rapid performance improvements. Across the HiRID cohorts (first, last, and 4093), AUROC increased to 0.812–0.861 with only 1\% of labeled data, and continued to rise toward near-saturated performance as more data were provided. Among these, the HiRID-last configuration consistently yielded the strongest zero-shot and fine-tuned performance, suggesting that events temporally closer to the prediction window carry higher prognostic value.

The P12 datasets, evaluated under 24-hour and 48-hour windows, showed more gradual but steady improvements with increasing fine-tuning data, converging to stable performance between the 50\% and 100\% data conditions. The eICU dataset exhibited similar trends, reflecting robust adaptation across a highly heterogeneous multi-institutional cohort.

Notably, across all external datasets except for eICU, PULSE-ICU fine-tuned with only 30\% of available target-domain data consistently outperformed models trained from random initialization using the full dataset. For eICU, however, the cohort size is substantially larger, which reduces the relative benefit of fine-tuning on top of pretrained representations and results in a more modest performance gain. On average, relative to zero-shot inference, AUROC improved by approximately +0.20 to +0.25 and AUPRC by +0.10 to +0.20, demonstrating the data efficiency and transferability of the PULSE-ICU foundation-style representations under diverse institutional and temporal conditions.

\FloatBarrier

\subsection{Comparison with existing models}
We compared the proposed PULSE-ICU model with previously published approaches across several representative tasks. All comparisons were made against the results reported in prior studies. Note that direct comparison across studies is inherently challenging due to differences in cohort construction, variable selection, input-window length, and embedding strategies. Moreover, comprehensive head-to-head comparison across all 18 downstream tasks is not feasible since most prior works typically report results on a limited subset of outcomes. Therefore, we focus our comparison on three tasks that are most widely evaluated in the literature—(i) in-hospital mortality, (ii) ICU mortality, and (iii) phenotype prediction—which allow consistent and meaningful benchmarking against existing models. Although direct comparison is infeasible, we report these results to contextualize the performance of our method within the broader literature on ICU representation learning.

\paragraph{In-hospital mortality}
Following the experimental setup described by Harutyunyan et al.~\cite{harutyunyan2019multitask}, prior works such as DuETT and ICU-BERT evaluated in-hospital mortality using 48-hour input windows, whereas our PULSE-ICU was fine-tuned using a 24-hour input window. For a conservative comparison, we evaluated our 24-hour–trained model directly on 48-hour inputs without additional retraining. As shown in Table~\ref{tab:inhospital_mortality_comparison}, our approach achieved an AUROC of 0.880 and an AUPRC of 0.597, which is slightly below DuETT but comparable to or better than other strong baselines including STraTS, Raindrop, and ICU-BERT. These findings indicate that the learned representations generalize effectively to longer temporal windows than those used during fine-tuning.

\paragraph{Phenotype prediction}
For phenotype prediction, prior ICU foundation models typically used representations derived from the entire ICU stay, making direct comparison with our 24-hour setting inappropriate. Therefore, we compared our PULSE-ICU with other 24-hour–based approaches, including CTPD~\cite{wang2025ctpd} and MedFuse~\cite{hayat2022medfuse}. As summarized in Table~\ref{tab:phenotype_comparison}, our model outperformed these approaches across both AUROC and AUPRC metrics, and even exceeded performance reported for multimodal EHR+imaging frameworks such as CTPD and MedFuse (EHR+CXR). These results suggest that the proposed embedding strategy captures clinically informative signals present within the early stages of ICU admission.

\begin{table*}[ht]
\centering
\caption{Comparison of in-hospital mortality prediction performance across prior models and the proposed PULSE-ICU. Results are reported as mean~$\pm$~standard deviation.}
\renewcommand{\arraystretch}{1.2}
\resizebox{0.4\textwidth}{!}{
\begin{tabular}{@{}lcc@{}}
\toprule
\textbf{Methods} & \textbf{AUROC} & \textbf{AUPRC} \\
\midrule
XGBoost & 0.886$\pm$0.002 & 0.593$\pm$0.004 \\
LSTM & 0.881$\pm$0.001 & 0.522$\pm$0.006 \\
mTAND & 0.864$\pm$0.002 & 0.540$\pm$0.007 \\
Raindrop & 0.878$\pm$0.001 & 0.546$\pm$0.002 \\
STraTS & 0.882$\pm$0.004 & 0.552$\pm$0.013 \\
DuETT & \textbf{0.912$\pm$0.020} & \textbf{0.627$\pm$0.002} \\
ICU-BERT & 0.875$\pm$0.007 & 0.558$\pm$0.014 \\
\midrule
PULSE-ICU (ours) & 0.880$\pm$0.006 & 0.597$\pm$0.011 \\
\bottomrule
\end{tabular}
}
\label{tab:inhospital_mortality_comparison}
\end{table*}

\begin{table*}[ht]
\centering
\caption{Comparison of phenotype prediction performance across prior models and the proposed PULSE-ICU. Results are reported as mean~$\pm$~standard deviation.}
\renewcommand{\arraystretch}{1.2}
\resizebox{0.5\textwidth}{!}{
\begin{tabular}{@{}lcc@{}}
\toprule
\textbf{Methods} & \textbf{AUROC} & \textbf{AUPRC} \\
\midrule
CNN & 0.675$\pm$0.014 & 0.370$\pm$0.002 \\
RNN & 0.669$\pm$0.049 & 0.363$\pm$0.028 \\
LSTM & 0.676$\pm$0.039 & 0.372$\pm$0.021 \\
Transformer & 0.674$\pm$0.024 & 0.373$\pm$0.031 \\
IP-Net & 0.682$\pm$0.020 & 0.384$\pm$0.027 \\
GRU-D & 0.505$\pm$0.034 & 0.231$\pm$0.036 \\
DGM-O & 0.597$\pm$0.034 & 0.299$\pm$0.025 \\
mTAND & 0.669$\pm$0.074 & 0.364$\pm$0.038 \\
SeFT & 0.571$\pm$0.004 & 0.269$\pm$0.009 \\
UTDE & 0.676$\pm$0.085 & 0.376$\pm$0.019 \\
CTPD & 0.719$\pm$0.040 & 0.424$\pm$0.060 \\
MedFuse (EHR) & 0.746$\pm$0.013 & 0.453$\pm$0.023 \\
MedFuse (EHR+CXR) & 0.770$\pm$0.013 & 0.481$\pm$0.024 \\
\midrule
PULSE-ICU (ours) & \textbf{0.785$\pm$0.002} & \textbf{0.509$\pm$0.003} \\
\bottomrule
\end{tabular}
}
\label{tab:phenotype_comparison}
\end{table*}

\paragraph{YAIB-aligned ICU mortality}
We further evaluated transfer performance under the YAIB variable schema using HiRID and eICU. Among the YAIB benchmark tasks, we compared performance on the ICU mortality prediction task. As shown in Table~\ref{tab:yaib comparison}, PULSE-ICU achieved the highest AUROC on the eICU dataset and the second-highest AUPRC, showing comparable or superior performance to all baseline methods. In the HiRID dataset, PULSE-ICU demonstrated a substantial improvement over existing approaches, achieving an AUROC of 0.927 and an AUPRC of 0.530. The particularly large gain in HiRID reflects the model's ability to capture long-range temporal dependencies and complex event interactions present in dense, high-frequency ICU sequences.

\paragraph{PhysioNet 2012 in-hospital mortality}
Following the common experimental setting in prior works, the P12 task was evaluated using a 48-hour observation window. On the P12 benchmark, PULSE-ICU demonstrated competitive performance relative to state-of-the-art irregular time-series methods, achieving an AUROC of 0.857 and an AUPRC of 0.544 (Table~\ref{tab:p12_comparison}). These results were comparable to the best-performing model, VITAL~\cite{kwon2025mind}, and exceeded the performance of several Transformer-, RNN-, and interpolation-based methods.

Collectively, across multiple tasks and datasets, the proposed model performed favorably relative to prior approaches, despite differences in input-window length and experimental design. These results highlight the versatility of the learned representations and their suitability for a wide range of ICU prediction tasks.

\begin{table*}[ht]
\centering
\caption{ICU mortality prediction performance on YAIB-aligned HiRID and eICU cohorts. For HiRID, the proposed model was evaluated under the HiRID-first configuration. Results are reported as mean~$\pm$~standard deviation.}
\renewcommand{\arraystretch}{1.2}
\resizebox{0.7\textwidth}{!}{
\begin{tabular}{@{}lcccc@{}}
\toprule
\multirow{2}{*}{\textbf{Methods}} &
\multicolumn{2}{c}{\textbf{HiRID}} &
\multicolumn{2}{c}{\textbf{eICU}} \\
\cmidrule(lr){2-3} \cmidrule(lr){4-5}
 & \textbf{AUROC} & \textbf{AUPRC} & \textbf{AUROC} & \textbf{AUPRC} \\ 
\midrule
\quad LR & 0.840$\pm$0.003 & 0.369$\pm$0.011 & 0.848$\pm$0.002 & 0.330$\pm$0.007 \\
\quad LGBM & 0.844$\pm$0.003 & 0.406$\pm$0.006 & 0.857$\pm$0.002 & \textbf{0.360$\pm$0.006} \\
\quad GRU & 0.848$\pm$0.002 & 0.394$\pm$0.004 & 0.860$\pm$0.001 & 0.356$\pm$0.001 \\
\quad LSTM & 0.840$\pm$0.007 & 0.378$\pm$0.010 & 0.855$\pm$0.002 & 0.357$\pm$0.004 \\
\quad TCN & 0.846$\pm$0.007 & 0.392$\pm$0.013 & 0.854$\pm$0.003 & 0.343$\pm$0.006 \\
\quad TF & 0.849$\pm$0.007 & 0.393$\pm$0.015 & 0.859$\pm$0.002 & 0.347$\pm$0.006 \\
\midrule
\quad PULSE-ICU (ours) & \textbf{0.927$\pm$0.002} & \textbf{0.530$\pm$0.012} & \textbf{0.864$\pm$0.001} & 0.359$\pm$0.003 \\
\bottomrule
\end{tabular}
}
\label{tab:yaib comparison}
\end{table*}

\begin{table*}[ht]
    \centering
    \caption{Mortality prediction performance on the PhysioNet 2012 (P12) dataset using a 48-hour observation window. Results are reported as mean~$\pm$~standard deviation.}
    \vspace{0.5em}
    \resizebox{0.5\textwidth}{!}{
    \begin{tabular}{@{}lcc@{}}
        \toprule
        \textbf{Methods} & \textbf{AUROC} & \textbf{AUPRC} \\ \midrule
        Transformer & 0.833 $\pm$ 0.007 & 0.479 $\pm$ 0.036 \\
        Trans-mean & 0.826 $\pm$ 0.020 & 0.463 $\pm$ 0.040 \\
        GRU-D & 0.819 $\pm$ 0.021 & 0.461 $\pm$ 0.047 \\
        SeFT & 0.739 $\pm$ 0.025 & 0.311 $\pm$ 0.041 \\
        mTAND & 0.842 $\pm$ 0.008 & 0.482 $\pm$ 0.034 \\
        IP-Net & 0.826 $\pm$ 0.014 & 0.476 $\pm$ 0.031 \\
        DGM$^2$-O & 0.834 $\pm$ 0.014 & 0.478 $\pm$ 0.033 \\
        MTGNN & 0.744 $\pm$ 0.067 & 0.355 $\pm$ 0.060 \\
        Raindrop & 0.828 $\pm$ 0.017 & 0.440 $\pm$ 0.030 \\
        ViTST & 0.851 $\pm$ 0.008 & 0.511 $\pm$ 0.041 \\ 
        VITAL & \textbf{0.860 $\pm$ 0.014} & \textbf{0.549 $\pm$ 0.042} \\ 
        \midrule
        PULSE-ICU (ours) & 0.857 $\pm$ 0.003 & 0.544 $\pm$ 0.003 \\
        \bottomrule
    \end{tabular}
    }
    \label{tab:p12_comparison}
\end{table*}

\subsection{Ablation study and representation analysis}

\paragraph{Embedding ablation}

To examine the contribution of individual embedding components, we performed a set of controlled ablation experiments by removing each component from the unified embedding module and fine-tuning the resulting model. Table~\ref{tab:ablation_delta} summarizes the performance change ($\Delta$) relative to the full PULSE-ICU model (Base).

Across all task groups, removing the value embedding produced the largest performance drop, particularly for multi-class SOFA prediction, indicating that continuous measurements carry essential signal that cannot be recovered through event-type or temporal information alone. This result demonstrates the effectiveness of the pretraining strategy, which explicitly incorporates value prediction as part of the learning objective. Removing the unit embedding also caused measurable degradation, suggesting that modeling the unit of measurement helps the model differentiate among heterogeneous physiological scales (e.g., arterial blood gases vs. serum chemistries). Excluding the time embedding yielded a moderate decline, confirming that explicit temporal encoding remains important even when the model receives irregular event timestamps as input.

Interestingly, removing the position embedding resulted in a small but consistent improvement. Because the model already incorporates precise absolute timing through the time embedding, the additional sequential position signal may introduce redundancy or noise in highly irregular ICU sequences. Another plausible explanation is that ICU event streams often contain multiple events sharing identical timestamps or occurring within very short intervals, which makes sequential indices only loosely correlated with true physiological progression. In such cases, learnable positional embeddings can exaggerate arbitrary ordering between near-simultaneous events, acting as a confounding signal rather than a helpful structural cue. This finding suggests that temporal alignment based on timestamps is more informative than traditional positional indexing for this data domain.

\begin{table*}[ht]
\centering
\caption{Ablation study results on the impact of attribute-specific embeddings in the unified event representation of PULSE-ICU. The Base column reports the average AUROC and AUPRC of the full model including all embedding components. Performance changes ($\Delta$) were measured by removing each embedding component (Value, Unit, Time, Position, Ordername, Orderdesc) and evaluating the model on a single validation fold under the same experimental configuration. A negative $\Delta$ indicates a performance degradation relative to the base model.
}
\vspace{0.3em}
\renewcommand{\arraystretch}{1.2}
\resizebox{\textwidth}{!}{
\begin{tabular}{@{}lccccccccccccccc@{}}
\toprule
\multirow{2}{*}{\textbf{Ablation Embedding}} 
& \multicolumn{2}{c}{\textbf{Base (Ref.)}} 
& \multicolumn{2}{c}{\textbf{Value ($\Delta$)}} 
& \multicolumn{2}{c}{\textbf{Unit ($\Delta$)}} 
& \multicolumn{2}{c}{\textbf{Time ($\Delta$)}} 
& \multicolumn{2}{c}{\textbf{Position ($\Delta$)}} 
& \multicolumn{2}{c}{\textbf{Ordername ($\Delta$)}} 
& \multicolumn{2}{c}{\textbf{Orderdesc ($\Delta$)}} \\ 
\cmidrule(lr){2-15}
& \textbf{AUROC} & \textbf{AUPRC}
& \textbf{AUROC} & \textbf{AUPRC}
& \textbf{AUROC} & \textbf{AUPRC}
& \textbf{AUROC} & \textbf{AUPRC}
& \textbf{AUROC} & \textbf{AUPRC}
& \textbf{AUROC} & \textbf{AUPRC}
& \textbf{AUROC} & \textbf{AUPRC} \\ 
\midrule
\textbf{Binary} 
& 0.865 & 0.454
& {$-$0.018} & {$-$0.030}
& {$-$0.010} & {$-$0.005}
& {$-$0.006} & {$-$0.001}
& {+0.009} & {+0.021}
& {$-$0.007} & {+0.001}
& {$-$0.004} & {+0.002} \\

\textbf{Multi-Class} 
& 0.899 & 0.585
& {$-$0.057} & {$-$0.105}
& {$-$0.017} & {$-$0.048}
& {$-$0.004} & {$-$0.008}
& {+0.007} & {+0.015}
& {$-$0.004} & {$-$0.005}
& {0.000} & {$-$0.002} \\

\textbf{Multi-Label} 
& 0.785 & 0.512
& {$-$0.021} & {$-$0.041}
& {$-$0.007} & {$-$0.015}
& {$-$0.003} & {$-$0.007}
& {+0.007} & {+0.006}
& {$-$0.003} & {$-$0.006}
& {$-$0.005} & {$-$0.012} \\

\textbf{All} 
& 0.872 & 0.501
& {$-$0.031} & {$-$0.056}
& {$-$0.012} & {$-$0.020}
& {$-$0.005} & {$-$0.003}
& {+0.008} & {+0.018}
& {$-$0.006} & {$-$0.001}
& {$-$0.003} & {0.000} \\
\bottomrule
\end{tabular}
}
\label{tab:ablation_delta}
\end{table*}

\begin{figure*}[ht]
    \centering
    \includegraphics[width=\textwidth]{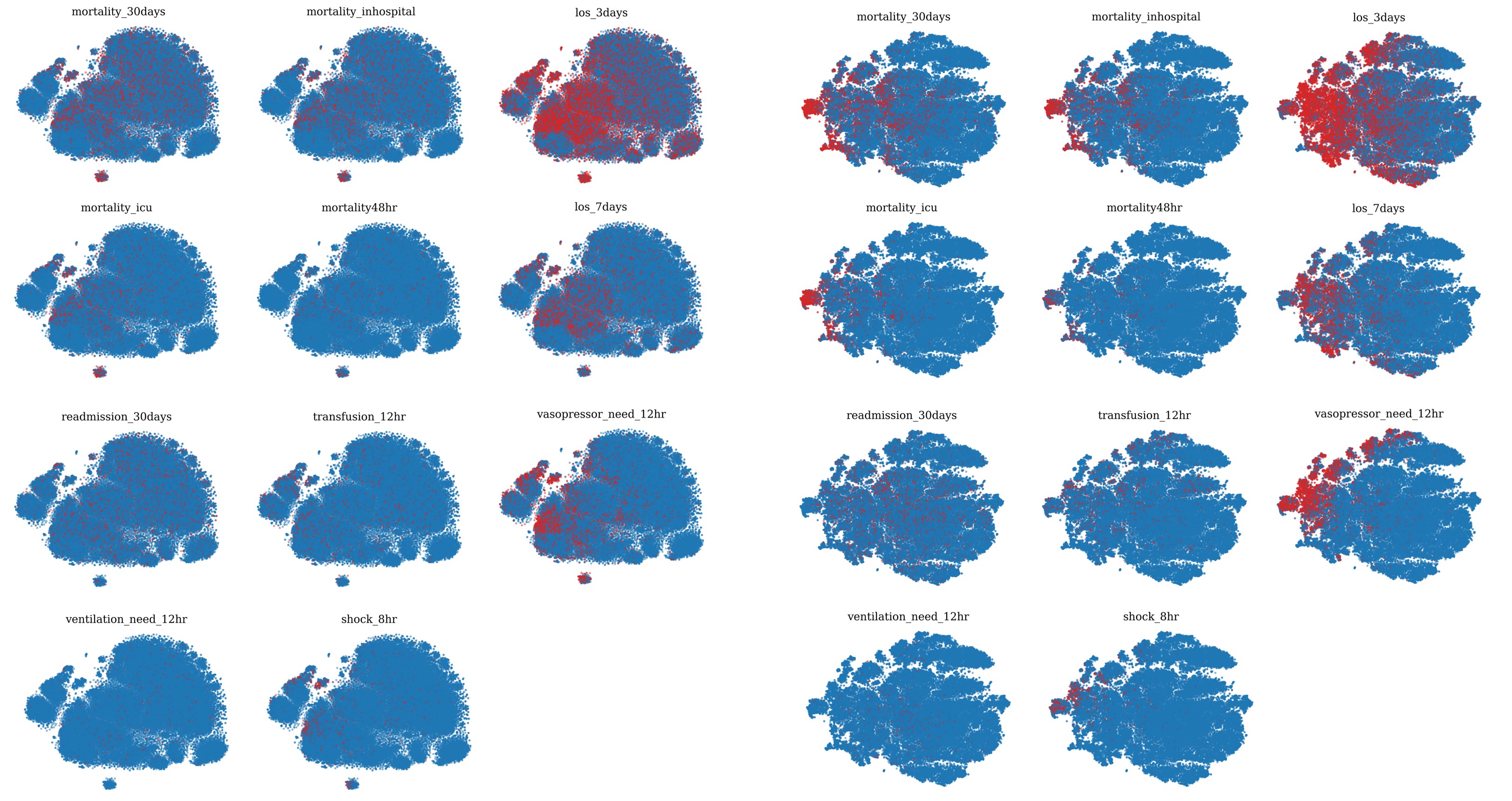}
    \caption{t-SNE visualization of patient-level latent representations for multiple binary prediction tasks in MIMIC-IV. Each point corresponds to an ICU stay and is colored by the binary outcome label (blue: negative, red: positive). The left column shows embeddings from a randomly initialized encoder, whereas the right column displays embeddings from the fine-tuned MIMIC model.
}
    \label{fig:embedding_visualization_mimic_binary}
\end{figure*}

\begin{figure*}[ht]
    \centering
    \includegraphics[width=0.8\textwidth]{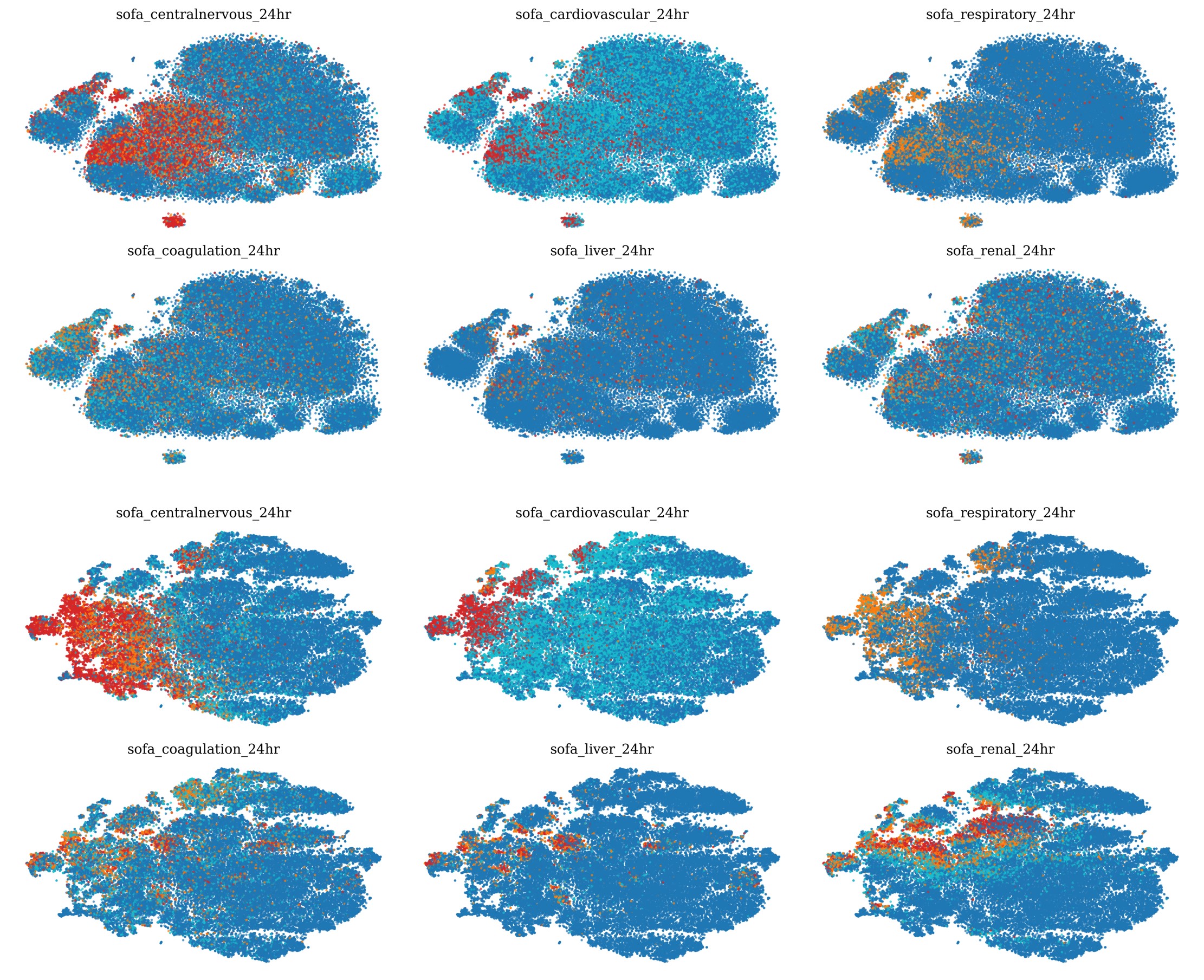}
    \caption{t-SNE visualization of patient-level latent representations for multi-class SOFA prediction tasks in MIMIC-IV. Each point represents an ICU stay and is colored by SOFA class, from low (blue) to high (red) severity. The top two rows show embeddings from a randomly initialized encoder, while the bottom two rows show embeddings from the fine-tuned model. Columns correspond to individual SOFA organ systems (central nervous, cardiovascular, respiratory, coagulation, liver, renal). 
}
    \label{fig:embedding_visualization_mimic_multiclass}
\end{figure*}

\paragraph{Representation visualization}
We further analyzed the learned representation space by projecting patient-level embeddings using t-SNE~\cite{maaten2008visualizing}. Each point corresponds to a patient’s unified representation and is colored according to task-specific labels. This allows for a direct examination of how patients with different clinical outcomes are organized within a common latent space. Figures~\ref{fig:embedding_visualization_mimic_binary} and \ref{fig:embedding_visualization_mimic_multiclass} compare embeddings obtained from a randomly initialized model and from the fine-tuned model, illustrating how fine-tuning reshapes the latent representation.

For binary prediction tasks, the fine-tuned representations exhibit clearer separation between positive and negative cases compared with the randomly initialized baseline, indicating that fine-tuning encourages the formation of clinically meaningful clusters. The effect is more pronounced for tasks with strong physiological signatures, such as short-term mortality and vasopressor requirements.

For multi-class SOFA tasks, the learned embeddings form structured gradients in the latent space, with higher SOFA classes distributed along a continuum rather than forming sharply separated clusters. This continuous organization aligns with the clinical nature of organ dysfunction, which progresses gradually rather than discretely. Fine-tuned embeddings also show reduced overlap across severity levels compared with the randomly initialized model, suggesting improved capture of organ-specific physiological patterns.

Taken together, the ablation and visualization findings demonstrate that (i) value- and unit-aware embeddings contribute substantially to predictive performance, (ii) timestamp-based temporal encoding is more informative than positional indexing in irregular clinical sequences, and (iii) the unified embedding space organizes patients along clinically interpretable axes that reflect disease severity and physiological deterioration.

\FloatBarrier

\FloatBarrier

\section{Discussion}

In this study, we developed PULSE-ICU, a clinical foundation model specialized for ICUs that jointly models heterogeneous clinical event streams and supports a broad set of downstream prediction tasks. Through self-supervised pretraining on large-scale ICU event sequences, PULSE-ICU learned contextual relationships among diverse clinical measurements by reconstructing masked events and values, enabling the extraction of temporally informed and semantically rich representations from irregular EHR data.

A key contribution of the framework is the unified embedding module that integrates event identity, continuous values, measurement units, and temporal attributes. This design allows the model to capture not only symbolic and categorical information but also the numerical and distributional characteristics of physiological variables, which are essential for clinical interpretation. The Longformer-based encoder further enables efficient modeling of long and densely sampled ICU trajectories, mitigating the computational constraints of conventional Transformer models while preserving global context.

Following pretraining, PULSE-ICU was fine-tuned on 18 clinically relevant prediction tasks, spanning mortality, length of stay, intervention forecasting, SOFA-based organ dysfunction classification, and phenotype identification. Joint fine-tuning yielded stable performance across task types, demonstrating that a shared representation space can support a wide range of clinical objectives. External validation on eICU, HiRID, and PhysioNet 2012 further showed that, while zero-shot performance reflected expected domain discrepancies, modest amounts of target-domain fine-tuning led to substantial improvements. These findings suggest that foundation-style pretraining can reduce the need for institution-specific model development and facilitate efficient adaptation to new hospital environments.

The model also demonstrated strong robustness in feature-limited scenarios. When restricted to a subset of core clinical variables, performance remained high, even with minimal fine-tuning, indicating that the model retains transferable structure even when applied to datasets with reduced variable coverage. Building on this adaptability, PULSE-ICU supports a total of 18 downstream clinical prediction tasks and, importantly, allows selective activation of individual tasks depending on the clinical objective. During external validation, different datasets were evaluated using different task subsets to reflect dataset-specific label availability. This confirms that the model can flexibly leverage task-specific classification heads based on data availability and application requirements. Collectively, PULSE-ICU offers a shared representation space that enables scalable and task-adaptive deployment, allowing practitioners to utilize only the relevant prediction tasks without requiring architectural modifications or retraining of the full model.

Despite these strengths, several limitations warrant consideration. First, the current framework focuses exclusively on structured clinical events. Many clinically salient signals—such as free-text notes, radiology images, and continuous waveform data—were not incorporated. Extending PULSE-ICU to a multimodal foundation framework, potentially via integration with large language models or time-series encoders for physiological waveforms, is a natural next step. Second, pretraining was conducted solely on MIMIC-IV; although external validation demonstrated strong adaptability, structural differences in coding systems, sampling frequencies, and value distributions across hospitals remain a major challenge. Approaches based on common data models or automated cross-dataset alignment may help better address these discrepancies. Third, the current implementation supports a fixed set of 18 downstream tasks. Scaling toward a more comprehensive clinical foundation model will require expanding the task set, accommodating dynamic task definitions, and developing flexible task heads that generalize to new objectives without reconfiguring the entire architecture.

In summary, this study presents PULSE-ICU, a self-supervised ICU foundation model that learns fine-grained representations of irregular clinical event sequences and supports scalable, multi-task clinical prediction. The results demonstrate strong within-domain performance, effective generalization under limited-variable and cross-institution settings, and substantial gains with minimal fine-tuning. As future work incorporates multimodal signals, harmonized event schemas, and an expanded task space, foundation-style modeling with PULSE-ICU has the potential to substantially enhance clinical decision support and improve the generalizability of predictive models across diverse ICU environments.

\section{Methods}

\begin{figure*} [ht]
    \centering
    \includegraphics[width=0.9\textwidth]{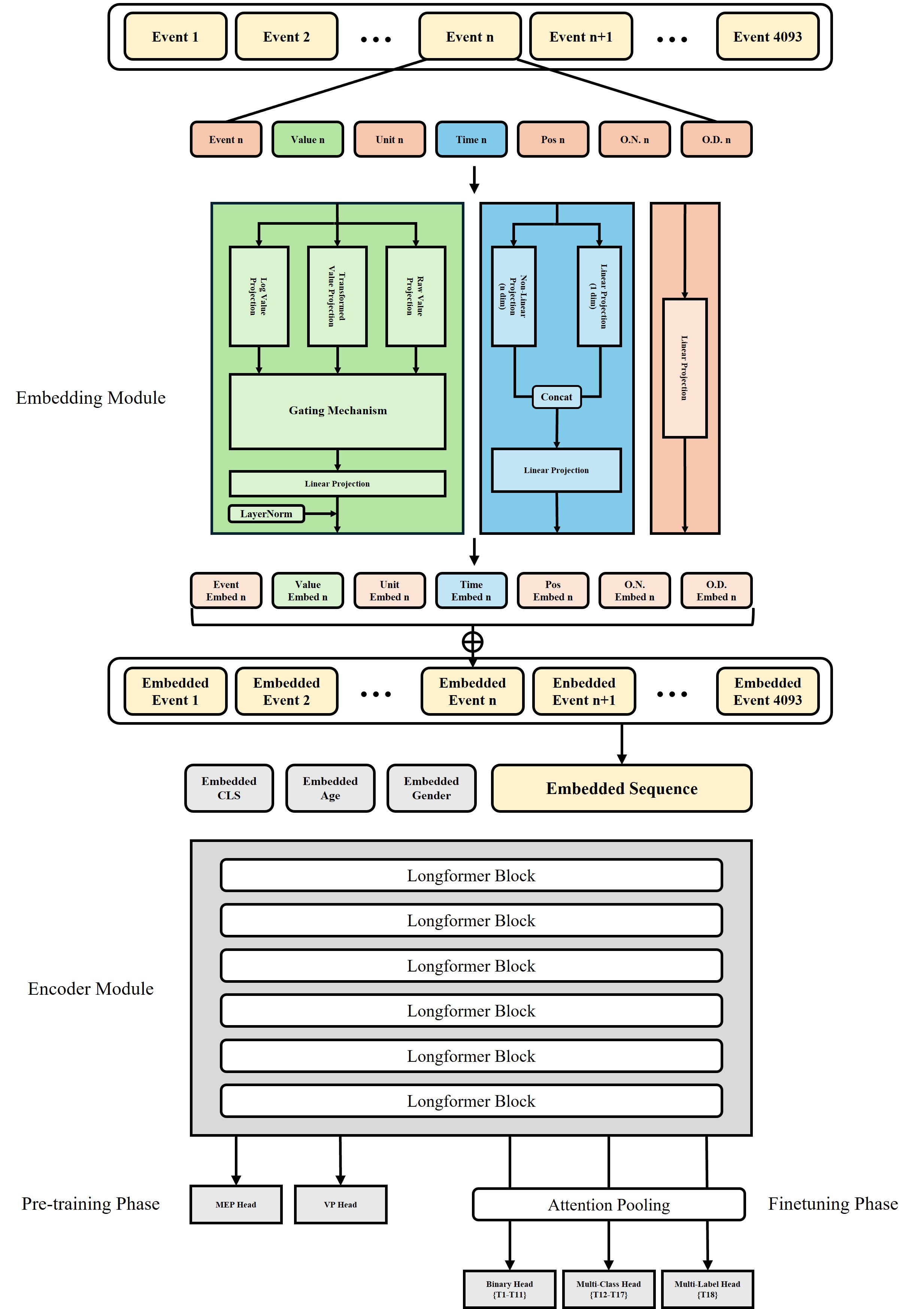}
    \caption{
Overview of the PULSE-ICU framework. The model consists of three main components: 
(1) an attribute-specific multi-embedding module that encodes heterogeneous ICU event attributes into unified token representations, 
(2) a Longformer-based encoder that captures long-range temporal dependencies in irregular, high-frequency clinical event streams, and 
(3) task-specific output heads used for both self-supervised pretraining (masked event and value prediction) and downstream multi-task fine-tuning across binary, multi-class, and multi-label clinical prediction tasks.
}
    \label{figure : model framework}
\end{figure*}

An overview of the proposed architectural framework, hereby referred to as PULSE-ICU, is provided in Figure~\ref{figure : model framework}. 
The model consists of three principal components: 
(i) an attribute-specific multi-embedding module that encodes heterogeneous ICU event attributes into a unified token representation, 
(ii) a Longformer-based encoder that captures long-range temporal dependencies inherent in irregular and high-frequency clinical event sequences, and 
(iii) task-specific output heads utilized during self-supervised pretraining and downstream multi-task fine-tuning. 
Each component is described in detail in the subsections that follow.

\FloatBarrier

\subsection{Datasets and cohorts}

We utilized the Medical Information Mart for Intensive Care IV (MIMIC-IV, version 2.2)~\cite{johnson2023mimic}, 
a large, de-identified EHR database collected at the Beth Israel Deaconess Medical Center 
between 2008 and 2019. MIMIC-IV contains more than 380,000 hospital admissions and approximately 70,000 
ICU stays, and includes structured data such as demographics, diagnoses, laboratory results, 
medications, vital signs, and procedures.
For this study, we extracted data from six core tables: \texttt{patients} (age, sex, and in-hospital mortality), 
\texttt{admissions} (hospital admission and discharge details), \texttt{icustays} (ICU admission and discharge times linked to hospital stays), \texttt{inputevents} (timestamped medication and fluid administration records, including dosage and administration route), \texttt{chartevents} (high-resolution clinical observations), and \texttt{procedureevents} (documented clinical procedures performed during ICU stays).

The study cohort included adult patients aged 18–100 years with ICU stays of at least 24 hours and a minimum of 10 recorded events. These criteria ensured sufficient temporal coverage and reduced noise from short or sparsely documented stays.

\subsection{Event sequence construction}

For each ICU stay, events from \texttt{chartevents}, \texttt{inputevents}, and \texttt{procedureevents} are merged into a single temporally ordered sequence. Data cleaning includes: (1) excluding physiologically implausible measurement values ($<$ 1st percentile or $>$ 99th percentile per variable), (2) removing implausible infusion rates ($>$ 99th percentile for each drug), (3) retaining only events occurring between ICU admission and discharge times. 

Each event was represented by seven attributes:
\begin{enumerate}
\item \textbf{Event} – clinical event name
\item \textbf{Value} – observed measurement (vital sign reading, laboratory value) or administered amount (medication dosage)
\item \textbf{Unit} – measurement unit
\item \textbf{Offset} – time from ICU admission (in days)
\item \textbf{Position} – sequential index within the stay (events with the same offset share the same position)
\item \textbf{Order name} – clinical category (e.g., fluids, invasive lines, enteral nutrition)
\item \textbf{Order description} – specific method or subtype (e.g., continuous infusion, bolus)
\end{enumerate}

The final representation for each ICU stay had a dimensionality of [$7 \times$ sequence length], with events sorted in chronological order. 

\subsection{Attribute-specific multi-embedding}

Each event attribute is tokenized and mapped into a continuous embedding space using attribute-specific embedding layers. This step transforms heterogeneous attribute types—categorical, numerical, and temporal—into a unified representation, enabling the model to jointly learn from both semantic and quantitative aspects of the data.

\paragraph{Categorical attributes}
Categorical attributes (\textit{event}, \textit{unit}, \textit{order name}, \textit{order description}) are embedded independently using separate learnable embedding layers. This separation allows the model to preserve domain-specific semantics within each attribute type. Static demographic variables (\textit{age}, \textit{gender}) are also treated as categorical tokens; they are embedded in the same manner and positioned at the beginning of the sequence to provide patient-level context for all downstream tasks.

\begin{align}
e_{\text{event}} &= W_{\text{event}}[e], &
e_{\text{unit}} &= W_{\text{unit}}[u], &
e_{\text{name}} &= W_{\text{name}}[n], \notag\\
e_{\text{desc}} &= W_{\text{desc}}[d], &
e_{\text{age}} &= W_{\text{age}}[a], &
e_{\text{gender}} &= W_{\text{gender}}[g].
\end{align}

\paragraph{Temporal and positional attributes}
The \textit{offset} attribute, representing elapsed time from ICU admission, is encoded using a Time2Vec~\cite{kazemi2019time2vec}-inspired module that combines a learnable linear term with sinusoidal terms to capture both absolute progression and periodic patterns. Specifically, for input time $t$, the embedding is computed as:
\begin{align}
\text{linear\_out} &= w_0 t + b_0, \\
\text{periodic\_out} &= \sin(Wt + b), \\
e_{\text{time}} &= W_t [\text{linear\_out}; \text{periodic\_out}] + b_t,
\end{align}
where $W,b \in \mathbb{R}^{F}$ denote the learnable frequency and phase parameters, and $W_t \in \mathbb{R}^{(F+1)\times d}$ projects the combined signal into the model dimension $d$.  
The \textit{position} attribute, denoting the sequential index of the event, is encoded via a learnable positional embedding to preserve event ordering in Transformer-based architectures~\cite{devlin2019bert}:
\begin{equation}
e_{\text{pos}} = W_{\text{pos}}[p].
\end{equation}

\paragraph{Continuous values}
ICU measurements exhibit diverse statistical characteristics. Some show narrow distributions, others are skewed, and some present long-tailed patterns. These distributional properties can vary greatly depending on hospitals, patient populations, and measurement protocols.
To ensure cross-dataset adaptability, we do not apply normalization or scaling to the value attributes. Instead, raw measurements are directly embedded through three complementary views:

\begin{itemize}
\item Raw view: linear projection of the original value to preserve magnitude.
\item Nonlinear view: transformation through a MLP with GELU activation to capture local variations and non-monotonic effects.
\item Log-scaled view: signed log transform with learnable scaling and a small offset followed by a linear projection to compress heavy tails while retaining directional information.
\end{itemize}

Formally, for a scalar value $v$, the three complementary views are defined as:
\begin{align}
e_{\text{raw}} &= W_{\text{raw}} v + b_{\text{raw}}, \\
e_{\text{nl}} &= W_{\text{nl},2} \, \text{GELU}(W_{\text{nl},1} v + b_{\text{nl},1}) + b_{\text{nl},2}, \\
v_{\log} &= \operatorname{sign}(v)\,\log(1 + |v| + \epsilon)\,s, \\
e_{\log} &= W_{\text{log}} v_{\log} + b_{\text{log}}.
\end{align}

These three representations are concatenated and passed through a gating network that generates three gate coefficients corresponding to each view. Specifically, the gating vector is computed as:
\begin{align}
\mathbf{g} &= [g_1, g_2, g_3] = \sigma(W_g [e_{\text{nl}}; e_{\text{raw}}; e_{\log}] + b_g),
\end{align}
where each scalar gate $g_i \in (0,1)$ adaptively modulates the contribution of the respective embedding. The final value embedding is then computed as:
\begin{align}
\mathbf{e}_{\text{value}} &= g_1 \odot e_{\text{nl}} + g_2 \odot e_{\text{raw}} + g_3 \odot e_{\log}, \\
\mathbf{e}_{\text{value}} &= \text{LayerNorm}(W_{\text{value}} \mathbf{e}_{\text{value}} + b_\text{value}).
\end{align}

This design explicitly models the interplay among raw magnitude, nonlinear transformation, and logarithmic compression. By learning the gate coefficients $(g_1, g_2, g_3)$ dynamically for each input, the model can balance scale sensitivity and nonlinear expressivity, improving robustness to extreme values while preserving clinically meaningful variations without requiring extensive per-variable preprocessing.

\paragraph{Integration of attribute embeddings}
For each event, the embeddings from the seven attributes—\textit{event}, \textit{offset}, \textit{position}, \textit{value}, \textit{unit}, \textit{order name}, and \textit{order description}—are summed element-wise to form the event representation:
\begin{equation}
e_{\text{Event}} =
e_{\text{event}}
+ e_{\text{time}}
+ e_{\text{pos}}
+ e_{\text{value}}
+ e_{\text{unit}}
+ e_{\text{name}}
+ e_{\text{desc}}.
\end{equation}

The sequence is chronologically ordered, with a special \texttt{[CLS]} token followed by static demographic tokens (\texttt{[AGE]}, \texttt{[GENDER]}) prepended to the beginning of the sequence.
Each event embedding $e_{\text{Event}, t} \in \mathbb{R}^{d_{\text{model}}}$ represents the $t$-th temporal event, and the sequence of all $L$ events is denoted as 
$e_{\text{Event}} = [e_{\text{Event}, 1}, \dots, e_{\text{Event}, L}] \in \mathbb{R}^{L \times d_{\text{model}}}$.

The final embedded sequence is constructed by concatenating static demographic tokens with the event sequence:
\begin{equation}
E_{\text{EHR}} =
\big[
e_{\text{cls}};
e_{\text{age}};
e_{\text{gender}};
e_{\text{Event}, 1:L}
\big]
\in \mathbb{R}^{(3 + L) \times d_{\text{model}}},
\end{equation}
where $L$ is the number of events and $d_{\text{model}}$ is the embedding dimension.

\subsection{Model architecture}
We adopt a Longformer~\cite{beltagy2020longformer} encoder to efficiently process long ICU event sequences. The encoder consists of 6 Transformer layers, each with 8 attention heads, $d_{\text{model}} = 512$, and feed-forward dimension $d_{\text{ff}} = 1024$.
Local sliding-window attention (width = 512) is applied to all tokens, while global attention is assigned to the \texttt{[CLS]} and demographic tokens to capture sequence-level and patient-level context. The maximum sequence length is fixed at 4{,}093 tokens, including special and padding tokens. For downstream classification, the sequence representation is obtained using a learnable scaled attention pooling mechanism~\cite{lin2017structured}.

\subsection{Pretraining}

Pretraining consists of two complementary tasks: Masked Event Prediction (MEP), which promotes contextual representation learning for discrete attributes, and Value Prediction (VP), which grounds embeddings in continuous measurement scales to enhance sensitivity to clinically meaningful variations.

For MEP, six of the seven attributes—\textit{event}, \textit{value}, \textit{unit}, \textit{offset}, \textit{order name}, and \textit{order description}—are masked, while \textit{position} is excluded. All six attributes of a selected event are masked simultaneously to prevent the model from inferring missing attributes from correlated ones. The model reconstructs the masked event attributes from surrounding context. To address class imbalance across event types (chartevents: 92.55\%, inputevents: 6.96\%, procedureevents: 0.49\%), we mask 30\% of events within each type, ensuring balanced learning across categories.

For events not masked in MEP, the value attribute of 11 clinically important variables (HR, RR, SaO\textsubscript{2}, ABPs, ABPd, Temperature, WBC, Sodium, Potassium, HCO\textsubscript{3}\textsuperscript{-}, Hemoglobin) is randomly masked with a probability of 5\%. The model predicts these masked values via regression, learning measurement scales and distributions to improve sensitivity to temporal fluctuations.

Pretraining data are partitioned at the ICU stay level into training and test sets. Stays exceeding 4,093 tokens in length are assigned exclusively to the training set and segmented into subsequences of length $\leq 4,093$. This strategy served two purposes: (i) ensuring that the test set consisted solely of complete ICU stays without artificial segmentation, thereby preserving evaluation fidelity, and (ii) maximizing the amount of pretraining data available by generating additional subsequences from long stays. The latter design choice allowed the model to learn from a broader range of temporal dependencies and event co-occurrence patterns. To further prevent data leakage, no individual ICU stay was included in both training and test sets. In total, the training set comprised 52,621 ICU stays (64,142 sequences), while the test set comprised 5,105 ICU stays (5,105 sequences).

Pretraining is conducted for 30 epochs using the AdamW optimizer with a learning rate of 1e-4 under a cosine annealing schedule. Training is performed in mixed-precision mode across multiple GPUs, and the best checkpoint is selected based on validation performance. This configuration ensures stable convergence and efficient optimization for long-sequence pretraining.

The tasks are jointly optimized via following loss functions:
\begin{align*}
\mathcal{L}_{\mathrm{MEP}} &= -\frac{1}{K} \sum_{k=1}^{K} \log \hat{y}_{k, t_k} \\
\mathcal{L}_{\mathrm{VP}} &= \frac{1}{K_v} \sum_{k=1}^{K_v} (\hat{v}_k - v_k)^2 \\
\mathcal{L}_{\mathrm{Total}} &= \mathcal{L}_{\mathrm{MEP}} + \lambda \cdot \mathcal{L}_{\mathrm{VP}}, \quad \lambda = 0.001
\end{align*}
where $K$ is the number of masked tokens in MEP and $K_v$ is the number of masked continuous values in VP. $\hat{y}_{k, t_k}$ is the predicted probability for the true token $t_k$ at position $k$, and $\hat{v}_k$/$v_k$ are predicted/true values. The weighting coefficient $\lambda$ balances the classification and regression losses.

\subsection{Multi-task fine-tuning}
\label{sec:fine-tuning}
We fine-tune the pretrained encoder in a multi-task setting across 18 clinically relevant downstream tasks, using a shared encoder and independent classification heads for each task.

\begin{itemize}
    \item \textbf{Binary classification} (11 tasks): prediction of mortality within 30 days of hospital admission, in-hospital mortality, ICU mortality, and mortality within 48 hours; prolonged length of stay (LOS; $>$3 days and $>$7 days); readmission within 30 days~\cite{harutyunyan2019multitask}; and early prediction of clinical interventions, including transfusion within 12~hours, vasopressor administration within 12~hours, and initiation of mechanical ventilation within 12~hours. In addition, we defined early prediction of shock within 8~hours as a distinct prognostic task. The shock label was assigned positive if, within the prediction window, a patient exhibited (i) lactate $\geq 2$ mmol/L and mean arterial pressure (MAP) $\leq 65$ mmHg, or (ii) administration of a vasopressor, following the established clinical definition of circulatory shock proposed by~\cite{singer2016third}.
    
    \item \textbf{Multi-class classification} (6 tasks): estimation of the Sequential Organ Failure Assessment (SOFA) score~\cite{ferreira2001serial} (0, 1, 2, $>=$3) for six organ systems (central nervous, cardiovascular, respiratory, coagulation, liver, and renal), enabling fine-grained assessment of organ dysfunction severity from early ICU data. The SOFA score was originally introduced by Vincent et al. for standardized evaluation of organ failure and outcome prediction in critically ill patients.
    
    \item \textbf{Multi-label classification} (1 task): prediction of 25 phenotype categories defined by the Healthcare Cost and Utilization Project Clinical Classifications Software (HCUP CCS)~\cite{hcup2012ccs}, which are also included in established ICU benchmarks~\cite{harutyunyan2019multitask}, capturing comorbidities and underlying conditions that may influence treatment strategies.
\end{itemize}

All tasks use a 24-hour observation window after ICU admission, followed by a 
12-hour prediction gap to prevent label leakage, and task-specific prediction windows. 
Each classification head consists of a dropout applied to the encoder representations, 
followed by a linear layer to produce predictions for its respective task. Binary and 
multi-label classification tasks are trained using binary cross-entropy loss, whereas 
multi-class classification tasks employ cross-entropy loss.

The total training objective is defined as the sum of loss functions over all downstream prediction tasks. 
Let $T_{\text{bin}}$, $T_{\text{mc}}$, and $T_{\text{ml}}$ denote the number of binary, multi-class, 
and multi-label tasks, respectively. In our implementation, we use $T_{\text{bin}} = 11$ binary tasks, 
$T_{\text{mc}} = 6$ multi-class tasks (each corresponding to a SOFA score component with $C_j = 6$ classes), 
and $T_{\text{ml}} = 1$ multi-label task for phenotype prediction consisting of $C_k = 25$ independent binary labels. 
The overall multi-task loss is formulated as:

\begin{align}
\mathcal{L}_{\text{Total}}
&=
\sum_{i=1}^{T_{\text{bin}}} 
\mathcal{L}^{(i)}_{\text{BCE}}
+
\sum_{j=1}^{T_{\text{mc}}}
\mathcal{L}^{(j)}_{\text{CE}}
+
\mathcal{L}^{(\text{ml})}_{\text{BCE}}
\\[8pt]
\mathcal{L}^{(i)}_{\text{BCE}}
&=
-\frac{1}{N}
\sum_{n=1}^{N}
\Big[
y^{(i)}_n \log \hat{y}^{(i)}_n
+
(1 - y^{(i)}_n)\log (1 - \hat{y}^{(i)}_n )
\Big]
\\[8pt]
\mathcal{L}^{(j)}_{\text{CE}}
&=
-\frac{1}{N}
\sum_{n=1}^{N}
\sum_{c=1}^{C_j}
y^{(j)}_{n,c}\log \hat{y}^{(j)}_{n,c}
\\[8pt]
\mathcal{L}^{(\text{ml})}_{\text{BCE}}
&=
-\frac{1}{N C_k}
\sum_{n=1}^{N}
\sum_{c=1}^{C_k}
\Big[
y^{(\text{ml})}_{n,c}\log \hat{y}^{(\text{ml})}_{n,c}
+
(1-y^{(\text{ml})}_{n,c})
\log (1-\hat{y}^{(\text{ml})}_{n,c})
\Big].
\end{align}

Model performance is evaluated using AUROC and AUPRC, with both 
macro- and micro-averaged metrics reported for multi-class and multi-label tasks.
Fine-tuning is performed for 30 epochs using the AdamW optimizer under a cosine annealing schedule, with separate learning rates of $5e^{-5}$ for the pretrained encoder and $2e^{-4}$ for the classification heads. The best checkpoint is selected based on validation performance.

\subsection{Experimental design}

We design a comprehensive experimental framework to evaluate the proposed ICU foundation-style model across diverse clinical conditions, data regimes, and institutional settings. Our evaluations are structured around two complementary 
aspects: (1) inference and transfer under limited variable availability, assessing the model’s robustness when only a subset of clinically essential features is accessible; and (2) external evaluation and domain adaptation across heterogeneous ICU databases, measuring zero-shot generalization and few-shot transferability to unseen clinical environments.

\subsubsection{Inference and transfer under limited variable availability}
\label{sec:limited_ver}

As described in Section~\ref{sec:fine-tuning}, the initial multi-task fine-tuning was conducted using all available variables from the MIMIC-IV dataset. However, in real-world clinical settings, the range of accessible variables may be restricted due to the structure of hospital information systems and data acquisition protocols. To assess whether predictive performance can be preserved under such constraints, we adopt the YAIB schema~\cite{van2023yet}, which defines a set of 52 variables commonly available across multiple ICU databases including MIMIC-IV, HiRID~\cite{hyland2020early}, eICU~\cite{pollard2018eicu}, and AmsterdamUMCDB~\cite{thoral2021sharing}.

Based on this schema, we reconstruct the input sequences using a total of 72 core variables extracted from MIMIC-IV, including vital signs, laboratory measurements, Glasgow Coma Scale features, vasopressor-related indicators, and mechanical ventilation status. The MIMIC-fine-tuned model is subsequently evaluated under this feature-restricted condition to analyze its robustness to limited variable availability.

To further evaluate the model’s adaptability under constrained data conditions and during domain transfer, we conduct experiments in both zero-shot and few-shot learning settings. Training data ratios are systematically varied across 0\%, 1\%, 5\%, 10\%, 30\%, 50\%, and 100\%. This experimental design enables a structured assessment of the model’s data efficiency, its ability to recover performance with limited supervision, and the adaptability of pretrained representations across clinical domains.

\subsubsection{External dataset evaluation and domain adaptation}

To assess the model's generalization and transferability to unseen clinical environments, we evaluate it using independent external datasets that were not included in either pretraining or fine-tuning: HiRID, eICU, and PhysioNet Challenge 2012 (P12)~\cite{silva2012predicting}. Evaluations focus on (i) zero-shot inference performance and (ii) the model's ability to adapt to new clinical domains. 

Given the structural differences between institutions, domain transfer experiments exclude MIMIC-specific embedding attributes (\textit{ordername} and \textit{orderdescription}) to ensure compatibility across databases. Furthermore, using the YAIB schema, the HiRID and eICU datasets are aligned in terms of variable definitions and temporal structure, enabling a consistent and reliable evaluation of cross-domain transfer.

The HiRID-YAIB cohort consists of 31,517 ICU stays, characterized by long, dense temporal sequences with an average of 3,471 events and 24 unique variables per stay. In contrast, the eICU-YAIB cohort comprises 123,764 ICU stays from multiple institutions, exhibiting a more sparse and heterogeneous temporal structure with an average of 215 events and 25 variables per stay.

\paragraph{HiRID-YAIB}
The High Time Resolution ICU Dataset (HiRID), collected from the ICUs of the University Hospital Bern, comprises approximately 33{,}000 ICU stays from 10{,}000 patients. It provides minute-level physiological signals, laboratory measurements, medication administrations, and interventions, enabling fine-grained temporal analysis. Given its dense temporal granularity and extended observation periods, HiRID serves as a benchmark for evaluating temporal generalization and long-range dependency modeling.

Since HiRID sequences are substantially longer than those in other ICU datasets, often exceeding the model’s maximum token capacity, we evaluate the first 24 hours after ICU admission using three complementary sequence configurations:\\
(i) using the first 4,093 tokens,\\
(ii) using the last 4,093 tokens, and\\
(iii) using the entire sequence when its length does not exceed 4,093 tokens.\\
This configuration enables systematic comparison of model performance across distinct temporal perspectives.

\paragraph{eICU-YAIB}
The eICU Collaborative Research Database (eICU-CRD) includes data from over 200 hospitals across the United States, encompassing more than 200,000 ICU admissions.
The dataset provides standardized EHR records covering demographics, vital signs, laboratory results, medications, and procedures, and captures substantial inter-hospital heterogeneity. 
We adopt the core variable schema from the YAIB framework to align variables between MIMIC-IV, HiRID, and eICU, enabling consistent feature representation across heterogeneous institutions.

\paragraph{P12}
The PhysioNet Challenge 2012 (P12) dataset comprises approximately 12,000 ICU admissions from multiple U.S. hospitals. Originally designed as a benchmark for mortality and severity-of-illness prediction, P12 contains low-frequency physiological and laboratory time-series data.
Although smaller in scale, its simplicity and uniform structure make it useful for assessing model transferability under low-data and low-resolution conditions. For P12, we evaluate two temporal input windows—the first 24 hours and the first 48 hours after ICU admission—to examine how varying the observation length influences predictive performance.
Since labels are defined within 48 hours, all prediction tasks are restricted to this time frame, and the mechanical ventilation task is specifically defined within the first 24-hour window.

\paragraph{Evaluation protocol}
Following the same preprocessing pipeline as MIMIC-IV, we first evaluate the multi-task fine-tuned model trained on MIMIC-IV in a zero-shot setting, where the model is directly applied to each external dataset without any additional training to assess pure domain generalization.
Subsequently, we perform domain adaptation experiments by further fine-tuning the same model on each external dataset. The proportion of training data is varied across 0\%, 1\%, 5\%, 10\%, 30\%, 50\%, and 100\%, and each model is fine-tuned for 10 epochs per setting using independent validation and test splits.
For the zero-shot evaluation, the MIMIC-IV fine-tuned model is evaluated using 5-fold cross-validation. In all domain adaptation settings, each experiment is repeated with three random seeds, and the average performance is reported.

Due to differences in data structure and label definitions, the set of prediction tasks that can be performed for each external dataset varies accordingly and is summarized in Table~\ref{table:external_tasks}. This enables the experimental design to flexibly adapt to the characteristics of each dataset, while still allowing for rigorous and comprehensive evaluation of generalization and transferability across heterogeneous clinical environments.

\begin{table*}[ht]
\centering
\caption{Downstream prediction task coverage across external evaluation datasets. 
A check mark (\checkmark) indicates tasks evaluated on each dataset.
Note that for the P12 dataset, since ICU LOS is not directly available, hospital LOS is used instead.}
\renewcommand{\arraystretch}{1.2}
\begin{tabular}{@{}lccc@{}}
\toprule
\textbf{Task} & \textbf{HiRID} & \textbf{eICU} & \textbf{P12} \\ \midrule
\multicolumn{4}{l}{\textbf{Binary}} \\
\quad Mortality 30days      & --   & --   & -- \\
\quad Mortality in Hospital      & --   & --   & \checkmark \\
\quad Mortality in ICU     & \checkmark & \checkmark & -- \\
\quad Mortality 48hr          & --   & \checkmark & -- \\ 
\quad ICU LOS 3 days      & \checkmark   & \checkmark   & \checkmark \\
\quad ICU LOS 7 days      & \checkmark   & \checkmark   & \checkmark \\ 
\quad Readmission 30days      & --   & --   & -- \\
\quad Transfusion 12hr & \checkmark & \checkmark & -- \\
\quad Vasopressor 12hr & \checkmark & \checkmark & -- \\
\quad Ventilation 12hr & \checkmark & \checkmark & \checkmark\textsuperscript{$\dagger$} \\
\quad Shock 8hr        & \checkmark & \checkmark & -- \\ \midrule
\multicolumn{4}{l}{\textbf{Multi-Class}} \\
\quad SOFA 24hr        & \checkmark   & \checkmark   & -- \\ \midrule
\multicolumn{4}{l}{\textbf{Multi-Label}} \\
\quad Phenotype         & --  & \checkmark   & -- \\ \bottomrule
\end{tabular}
\\[3pt]
{\footnotesize $^{\dagger}$Evaluated only in the 24-hour setup, as P12 provides a single 48-hour input sequence.}
\label{table:external_tasks}
\end{table*}

\section{Appendix}

\subsection{Temporal generalization across observation windows}
\label{sec:temporal-generalization}

\begin{figure*}[ht]
    \centering
    \includegraphics[width=0.9\textwidth]{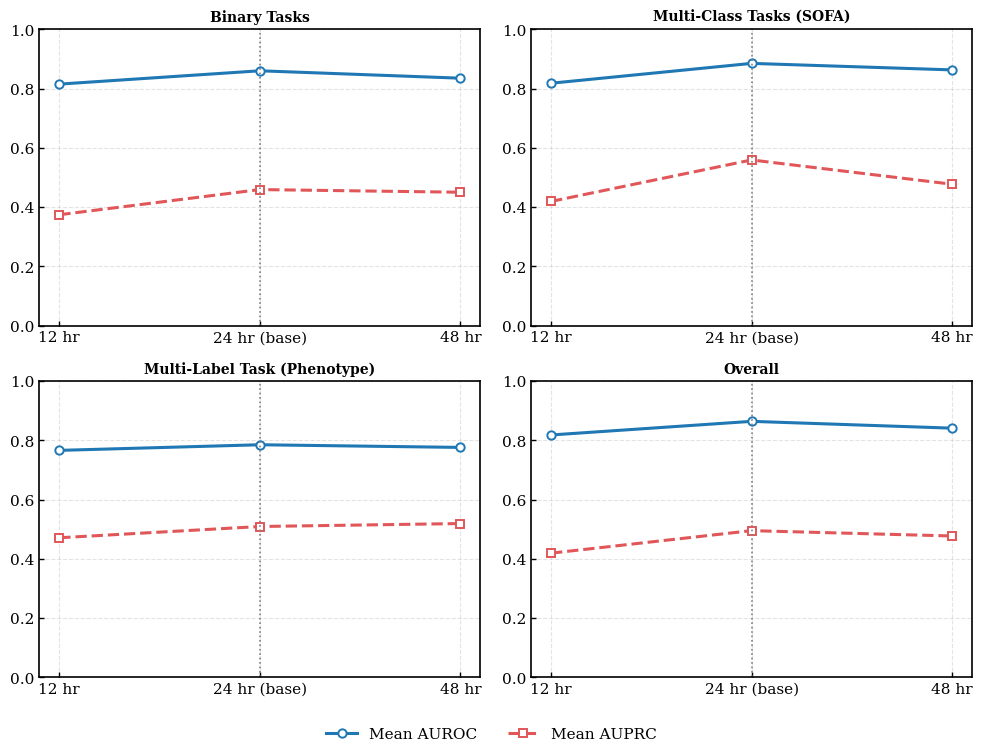}
    \caption{
Temporal generalization performance of the 24-hour fine-tuned model across different observation windows.
The model fine-tuned using 24-hour sequences is evaluated on datasets constructed with 12-hour and 48-hour observation windows without additional training.
Both AUROC and AUPRC are reported for each task category: Binary, Multi-Class (SOFA), Multi-Label (Phenotype), and Overall mean performance.
}
    \label{figure : Observation Window}
\end{figure*}

\begin{figure*}[ht]
    \centering
    \includegraphics[width=0.9\textwidth]{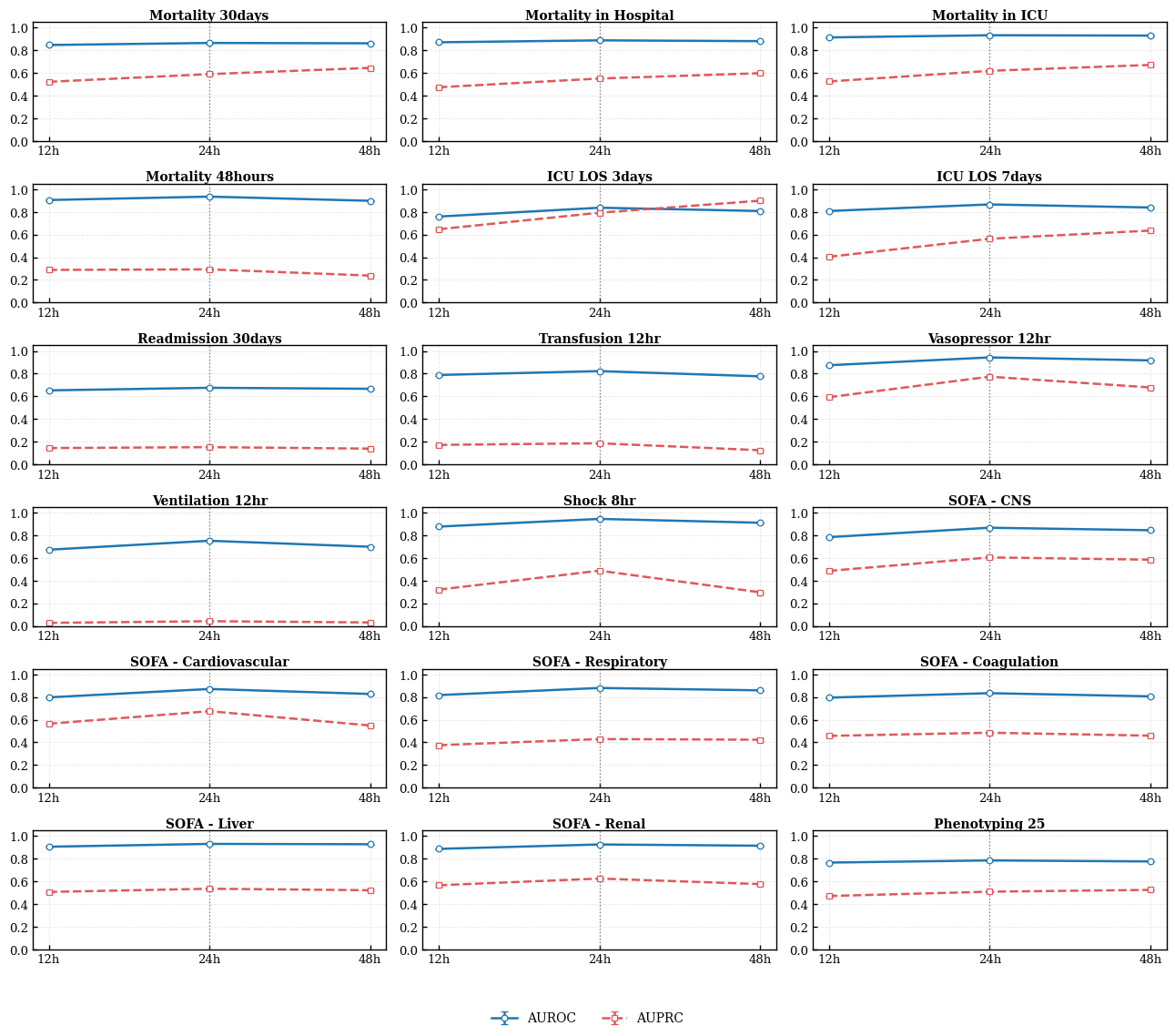}
    \caption{
Temporal transfer performance of the 24-hour fine-tuned model across different observation windows (12h, 48h).
Each subplot represents the AUROC and AUPRC scores of a specific binary, multi-class (SOFA), or multi-label prediction task.
The model, fine-tuned using 24-hour sequences, was evaluated without additional training on 12-hour and 48-hour versions of the same dataset.
}
    \label{figure : Observation Window All Tasks}
\end{figure*}

This section evaluates the temporal generalization ability of the PULSE-ICU model fine-tuned using 24-hour observation windows. Specifically, the 24-hour fine-tuned model is tested on datasets constructed with 12-hour and 48-hour observation windows, without any additional training, to assess whether it maintains predictive accuracy under different temporal contexts. 
Figure~\ref{figure : Observation Window} summarizes performance trends across binary, multi-class, and multi-label task groups, as well as the overall mean performance, in terms of both AUROC and AUPRC. Per-task results are provided in Figure~\ref{figure : Observation Window All Tasks}.

Overall, the model exhibited comparable performance when evaluated on 12-hour and 48-hour inputs, 
closely matching its performance under the original 24-hour configuration. This consistency indicates strong temporal adaptability and suggests that the model effectively captures clinically relevant temporal dynamics across varying observation-window lengths. 
In practice, the proposed model thus generalizes robustly across shorter and longer early-stage input windows while maintaining stable predictive performance.

\FloatBarrier

\begin{table*}[ht]
\centering
\caption{Label distributions across clinical prediction tasks in the MIMIC-IV dataset under 12-hour, 24-hour, and 48-hour input windows. Values indicate the proportion of each class.}
\renewcommand{\arraystretch}{1.2}
\resizebox{\textwidth}{!}{
\begin{tabular}{@{}llccc>{\centering\arraybackslash}p{0.05\textwidth}ccc>{\centering\arraybackslash}p{0.05\textwidth}ccc>{\centering\arraybackslash}p{0.05\textwidth}@{}}
\toprule
\multirow{2}{*}{\textbf{Task Type}} & \multirow{2}{*}{\textbf{Task Name}} 
& \multicolumn{4}{c}{\textbf{12-hour Window}} 
& \multicolumn{4}{c}{\textbf{24-hour Window}} 
& \multicolumn{4}{c}{\textbf{48-hour Window}} \\ 
\cmidrule(lr){3-6} \cmidrule(lr){7-10} \cmidrule(lr){11-14}
 &  & \textbf{0} & \textbf{1} & \textbf{2} & \textbf{$\geq3$} 
 & \textbf{0} & \textbf{1} & \textbf{2} & \textbf{$\geq3$} 
 & \textbf{0} & \textbf{1} & \textbf{2} & \textbf{$\geq3$} \\ 
\midrule
\textbf{Binary} 
& Mortality 30days & 0.86 & 0.14 &  &  & 0.85 & 0.15 &  &  & 0.84 & 0.16 &  &  \\
& Mortality in Hospital & 0.91 & 0.09 &  &  & 0.90 & 0.10 &  &  & 0.89 & 0.11 &  &  \\
& Mortality in ICU & 0.94 & 0.06 &  &  & 0.94 & 0.06 &  &  & 0.93 & 0.07 &  &  \\
& Mortality 48hr & 0.97 & 0.03 &  &  & 0.98 & 0.02 &  &  & 0.98 & 0.02 &  &  \\
& ICU LOS 3days & 0.66 & 0.34 &  &  & 0.59 & 0.41 &  &  & 0.56 & 0.44 &  &  \\
& ICU LOS 7days & 0.88 & 0.12 &  &  & 0.86 & 0.14 &  &  & 0.85 & 0.15 &  &  \\
& Readmission 30days & 0.92 & 0.08 &  &  & 0.92 & 0.08 &  &  & 0.91 & 0.09 &  &  \\
& Transfusion 12hr & 0.95 & 0.05 &  &  & 0.96 & 0.04 &  &  & 0.95 & 0.05 &  &  \\
& Vasopressor 12hr & 0.86 & 0.14 &  &  & 0.87 & 0.13 &  &  & 0.86 & 0.14 &  &  \\
& Ventilation 12hr & 0.99 & 0.01 &  &  & 0.99 & 0.01 &  &  & 0.99 & 0.01 &  &  \\
& Shock 8hr & 0.97 & 0.03 &  &  & 0.97 & 0.03 &  &  & 0.97 & 0.03 &  &  \\
\midrule
\textbf{Multi-Label}
& Acute and unspecified renal failure & 0.70 & 0.30 &  &  & 0.68 & 0.32 &  &  & 0.67 & 0.33 &  &  \\
\textbf{(Phenotype)}
& Acute cerebrovascular disease & 0.91 & 0.09 &  &  & 0.90 & 0.10 &  &  & 0.90 & 0.10 &  &  \\
& Acute myocardial infarction & 0.92 & 0.08 &  &  & 0.91 & 0.09 &  &  & 0.91 & 0.09 &  &  \\
& Cardiac dysrhythmias & 0.61 & 0.39 &  &  & 0.59 & 0.41 &  &  & 0.59 & 0.41 &  &  \\
& Chronic kidney disease & 0.78 & 0.22 &  &  & 0.78 & 0.22 &  &  & 0.78 & 0.22 &  &  \\
& COPD and bronchiectasis & 0.86 & 0.14 &  &  & 0.86 & 0.14 &  &  & 0.86 & 0.14 &  &  \\
& Complications of surgery/medical care & 0.76 & 0.24 &  &  & 0.74 & 0.26 &  &  & 0.74 & 0.26 &  &  \\
& Conduction disorders & 0.89 & 0.11 &  &  & 0.89 & 0.11 &  &  & 0.88 & 0.12 &  &  \\
& CHF; nonhypertensive & 0.73 & 0.27 &  &  & 0.71 & 0.29 &  &  & 0.70 & 0.30 &  &  \\
& Coronary atherosclerosis & 0.71 & 0.29 &  &  & 0.70 & 0.30 &  &  & 0.70 & 0.30 &  &  \\
& Diabetes w/ complications & 0.86 & 0.14 &  &  & 0.86 & 0.14 &  &  & 0.86 & 0.14 &  &  \\
& Diabetes w/o complication & 0.82 & 0.18 &  &  & 0.81 & 0.19 &  &  & 0.81 & 0.19 &  &  \\
& Lipid metabolism disorders & 0.59 & 0.41 &  &  & 0.58 & 0.42 &  &  & 0.58 & 0.42 &  &  \\
& Essential hypertension & 0.59 & 0.41 &  &  & 0.59 & 0.41 &  &  & 0.59 & 0.41 &  &  \\
& Fluid/electrolyte disorders & 0.58 & 0.42 &  &  & 0.57 & 0.43 &  &  & 0.57 & 0.43 &  &  \\
& GI hemorrhage & 0.92 & 0.08 &  &  & 0.92 & 0.08 &  &  & 0.92 & 0.08 &  &  \\
& Hypertension w/ complications & 0.77 & 0.23 &  &  & 0.77 & 0.23 &  &  & 0.77 & 0.23 &  &  \\
& Other liver diseases & 0.84 & 0.16 &  &  & 0.84 & 0.16 &  &  & 0.84 & 0.16 &  &  \\
& Other lower respiratory disease & 0.89 & 0.11 &  &  & 0.88 & 0.12 &  &  & 0.88 & 0.12 &  &  \\
& Other upper respiratory disease & 0.94 & 0.06 &  &  & 0.94 & 0.06 &  &  & 0.94 & 0.06 &  &  \\
& Pleurisy / pneumothorax / collapse & 0.88 & 0.12 &  &  & 0.87 & 0.13 &  &  & 0.87 & 0.13 &  &  \\
& Pneumonia (non-TB) & 0.84 & 0.16 &  &  & 0.83 & 0.17 &  &  & 0.83 & 0.17 &  &  \\
& Respiratory failure / arrest & 0.71 & 0.29 &  &  & 0.68 & 0.32 &  &  & 0.68 & 0.32 &  &  \\
& Septicemia (non-labor) & 0.81 & 0.19 &  &  & 0.79 & 0.21 &  &  & 0.79 & 0.21 &  &  \\
& Shock & 0.89 & 0.11 &  &  & 0.88 & 0.12 &  &  & 0.88 & 0.12 &  &  \\
\midrule
\textbf{Multi-Class} 
& Central Nervous System & 0.61 & 0.15 & 0.08 & 0.16 & 0.61 & 0.14 & 0.08 & 0.16 & 0.61 & 0.14 & 0.09 & 0.16 \\
\textbf{(SOFA)} 
& Cardiovascular & 0.44 & 0.46 & 0.01 & 0.09 & 0.48 & 0.42 & 0.01 & 0.09 & 0.49 & 0.41 & 0.01 & 0.09 \\ 
& Respiratory & 0.89 & 0.02 & 0.09 & 0.01 & 0.89 & 0.02 & 0.09 & 0.01 & 0.88 & 0.02 & 0.09 & 0.01 \\
& Coagulation & 0.76 & 0.14 & 0.08 & 0.03 & 0.76 & 0.13 & 0.08 & 0.03 & 0.76 & 0.13 & 0.08 & 0.03 \\
& Liver & 0.92 & 0.03 & 0.03 & 0.02 & 0.93 & 0.03 & 0.03 & 0.02 & 0.93 & 0.03 & 0.03 & 0.02 \\
& Renal & 0.75 & 0.13 & 0.07 & 0.05 & 0.77 & 0.12 & 0.07 & 0.04 & 0.77 & 0.12 & 0.07 & 0.04 \\
\bottomrule
\end{tabular}}
\label{tab:mimic_label_distribution}
\end{table*}

\begin{table*}[ht]
\centering
\caption{Label distribution across clinical prediction tasks in the eICU dataset (24-hour input window).}
\renewcommand{\arraystretch}{1.2}
\resizebox{\textwidth}{!}{
\begin{tabular}{@{}llccc>{\centering\arraybackslash}p{0.05\textwidth}ccc>{\centering\arraybackslash}p{0.05\textwidth}ccc>{\centering\arraybackslash}p{0.05\textwidth}@{}}
\toprule
\textbf{Task Type} & \textbf{Task Name} & \textbf{0} & \textbf{1} & \textbf{2} & \textbf{$\geq3$} \\
\midrule
\textbf{Binary} 
& Mortality in ICU & 0.946 & 0.054 &  &  \\
& Mortality 48hr & 0.977 & 0.023 &  &  \\
& ICU LOS 3days & 0.622 & 0.378 &  &  \\
& ICU LOS 7days & 0.884 & 0.116 &  &  \\
& Readmission 30days & 0.912 & 0.088 &  &  \\
& Transfusion 12hr & 0.982 & 0.018 &  &  \\
& Vasopressor 12hr & 0.925 & 0.075 &  &  \\
& Ventilation 12hr & 0.999 & 0.001 &  &  \\
& Shock 8hr & 0.990 & 0.010 &  &  \\
\midrule
\textbf{Multi-Label}
& Acute and unspecified renal failure & 0.878 & 0.122 &  &  \\
\textbf{(Phenotype)}
& Acute cerebrovascular disease & 0.927 & 0.073 &  &  \\
& Acute myocardial infarction & 0.944 & 0.056 &  &  \\
& Cardiac dysrhythmias & 0.866 & 0.134 &  &  \\
& Chronic kidney disease & 0.908 & 0.092 &  &  \\
& COPD and bronchiectasis & 0.917 & 0.083 &  &  \\
& Complications of surgery / medical care & 0.990 & 0.010 &  &  \\
& Conduction disorders & 0.992 & 0.008 &  &  \\
& CHF; nonhypertensive & 0.899 & 0.101 &  &  \\
& Coronary atherosclerosis & 0.972 & 0.028 &  &  \\
& Diabetes with complications & 0.889 & 0.111 &  &  \\
& Diabetes without complication & 0.864 & 0.136 &  &  \\
& Fluid / electrolyte disorders & 0.909 & 0.091 &  &  \\
& GI hemorrhage & 0.986 & 0.014 &  &  \\
& Hypertension with complications & 0.948 & 0.052 &  &  \\
& Lipid metabolism disorders & 0.917 & 0.083 &  &  \\
& Other liver diseases & 0.974 & 0.026 &  &  \\
& Other lower respiratory disease & 0.977 & 0.023 &  &  \\
& Other upper respiratory disease & 0.989 & 0.011 &  &  \\
& Pleurisy / pneumothorax / collapse & 0.979 & 0.021 &  &  \\
& Pneumonia (non-TB) & 0.905 & 0.095 &  &  \\
& Respiratory failure / arrest & 0.901 & 0.099 &  &  \\
& Septicemia (non-labor) & 0.868 & 0.132 &  &  \\
& Shock & 0.936 & 0.064 &  &  \\

\midrule
\textbf{Multi-Class}
& Central Nervous System & 0.713 & 0.089 & 0.058 & 0.140 \\
\textbf{(SOFA)} 
& Cardiovascular & 0.932 & 0.054 & 0.009 & 0.005 \\
& Respiratory & 0.881 & 0.019 & 0.100 & 0.002 \\
& Coagulation & 0.799 & 0.117 & 0.067 & 0.017 \\
& Liver & 0.949 & 0.024 & 0.021 & 0.007 \\
& Renal & 0.794 & 0.106 & 0.060 & 0.041 \\
\bottomrule
\end{tabular}
}
\label{tab:eicu_label_distribution}
\end{table*}

\begin{table*}[ht]
\centering
\caption{Label distribution across clinical prediction tasks in the HiRID dataset (24-hour input window). Values indicate the proportion of each class for three sequence versions (first, last, and 4093).}
\renewcommand{\arraystretch}{1.2}
\resizebox{\textwidth}{!}{
\begin{tabular}{@{}llcccccccccccc@{}}
\toprule
\textbf{Task Type} & \textbf{Task Name} &
\multicolumn{4}{c}{\textbf{first ver.}} &
\multicolumn{4}{c}{\textbf{last ver.}} &
\multicolumn{4}{c}{\textbf{4093 ver.}} \\
\cmidrule(lr){3-6} \cmidrule(lr){7-10} \cmidrule(lr){11-14}
 & & \textbf{0} & \textbf{1} & \textbf{2} & \textbf{$\geq3$} & 
   \textbf{0} & \textbf{1} & \textbf{2} & \textbf{$\geq3$} &
   \textbf{0} & \textbf{1} & \textbf{2} & \textbf{$\geq3$} \\
\midrule
\textbf{Binary} 
& Mortality in ICU & 0.943 & 0.057 &  &  & 0.943 & 0.057 &  &  & 0.964 & 0.036 &  &  \\
& ICU LOS 3days & 0.815 & 0.185 &  &  & 0.815 & 0.185 &  &  & 0.910 & 0.091 &  &  \\
& ICU LOS 7days & 0.929 & 0.071 &  &  & 0.929 & 0.071 &  &  & 0.971 & 0.029 &  &  \\
& Transfusion 12hr & 0.974 & 0.026 &  &  & 0.974 & 0.026 &  &  & 0.988 & 0.012 &  &  \\
& Vasopressor 12hr & 0.916 & 0.084 &  &  & 0.916 & 0.084 &  &  & 0.974 & 0.026 &  &  \\
& Ventilation 12hr & 0.831 & 0.169 &  &  & 0.831 & 0.169 &  &  & 0.966 & 0.034 &  &  \\
& Shock 8hr & 0.971 & 0.029 &  &  & 0.971 & 0.029 &  &  & 0.992 & 0.008 &  &  \\
\midrule
\textbf{Multi-Class}
& Central Nervous System & 0.742 & 0.136 & 0.042 & 0.080 & 0.742 & 0.136 & 0.042 & 0.080 & 0.856 & 0.095 & 0.025 & 0.023 \\
\textbf{(SOFA)} 
& Cardiovascular & 0.707 & 0.204 & 0.021 & 0.068 & 0.707 & 0.204 & 0.021 & 0.068 & 0.851 & 0.120 & 0.004 & 0.026 \\
& Respiratory & 0.862 & 0.015 & 0.063 & 0.060 & 0.862 & 0.015 & 0.063 & 0.060 & 0.974 & 0.004 & 0.011 & 0.011 \\
& Coagulation & 0.866 & 0.064 & 0.050 & 0.020 & 0.866 & 0.064 & 0.050 & 0.020 & 0.946 & 0.030 & 0.017 & 0.008 \\
& Liver & 0.966 & 0.012 & 0.016 & 0.006 & 0.966 & 0.012 & 0.016 & 0.006 & 0.986 & 0.005 & 0.006 & 0.003 \\
& Renal & 0.920 & 0.039 & 0.029 & 0.013 & 0.920 & 0.039 & 0.029 & 0.013 & 0.970 & 0.015 & 0.010 & 0.005 \\
\bottomrule
\end{tabular}
}
\label{tab:hirid_label_distribution}
\end{table*}

\begin{table*}[ht]
\centering
\caption{Label distribution across binary clinical prediction tasks in the P12 dataset under 24-hour and 48-hour input windows.}
\renewcommand{\arraystretch}{1.2}
\resizebox{\textwidth}{!}{
\begin{tabular}{@{}lcccc@{}}
\toprule
\multirow{2}{*}{\textbf{Task Name}} & 
\multicolumn{2}{c}{\textbf{24h}} & 
\multicolumn{2}{c}{\textbf{48h}} \\ 
\cmidrule(lr){2-3} \cmidrule(lr){4-5}
 & \textbf{0} & \textbf{1} & \textbf{0} & \textbf{1} \\
\midrule
Mortality in Hospital & 0.858 & 0.142 & 0.855 & 0.145 \\
LOS 3days & 0.027 & 0.973 & 0.030 & 0.970 \\
LOS 7days & 0.281 & 0.719 & 0.282 & 0.718 \\
Ventilation 12hr & 0.505 & 0.495 & - & - \\
\bottomrule
\end{tabular}
}
\label{tab:p12_label_distribution}
\end{table*}

\begin{figure*}[ht]
    \centering
    \includegraphics[width=0.9\textwidth]{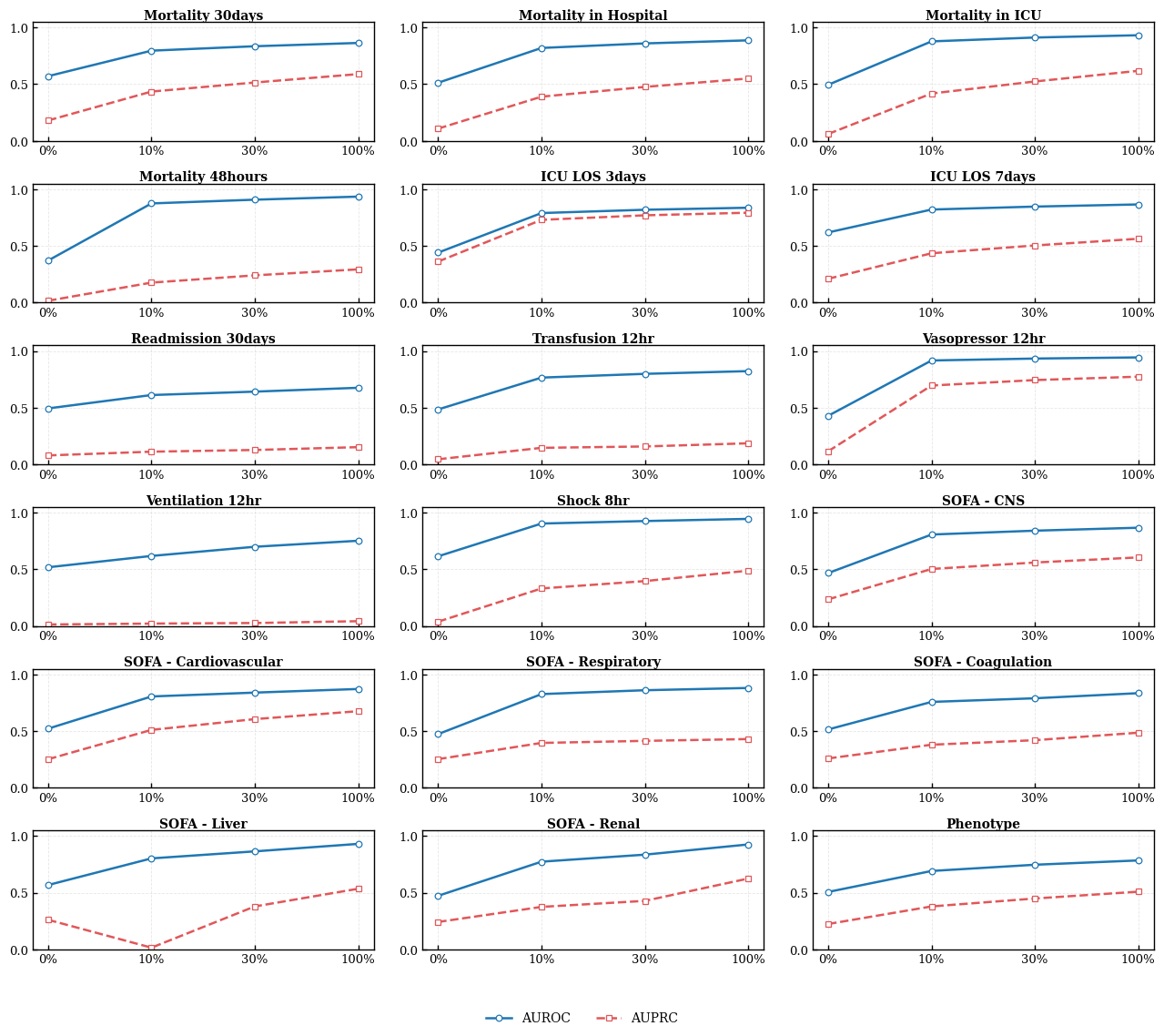}
    \caption{
Performance comparison across all clinical prediction tasks under varying fine-tuning data ratios.
Each subplot shows AUROC and AUPRC for individual binary, multi-class (SOFA), and multi-label tasks as the fine-tuning data ratio increases from 0\% (zero-shot) to 100\%.}
    \label{figure : Main Finetuning Ratio All Tasks}
\end{figure*}

\FloatBarrier

\FloatBarrier

\begin{figure*}[ht]
    \centering
    \includegraphics[width=0.9\textwidth]{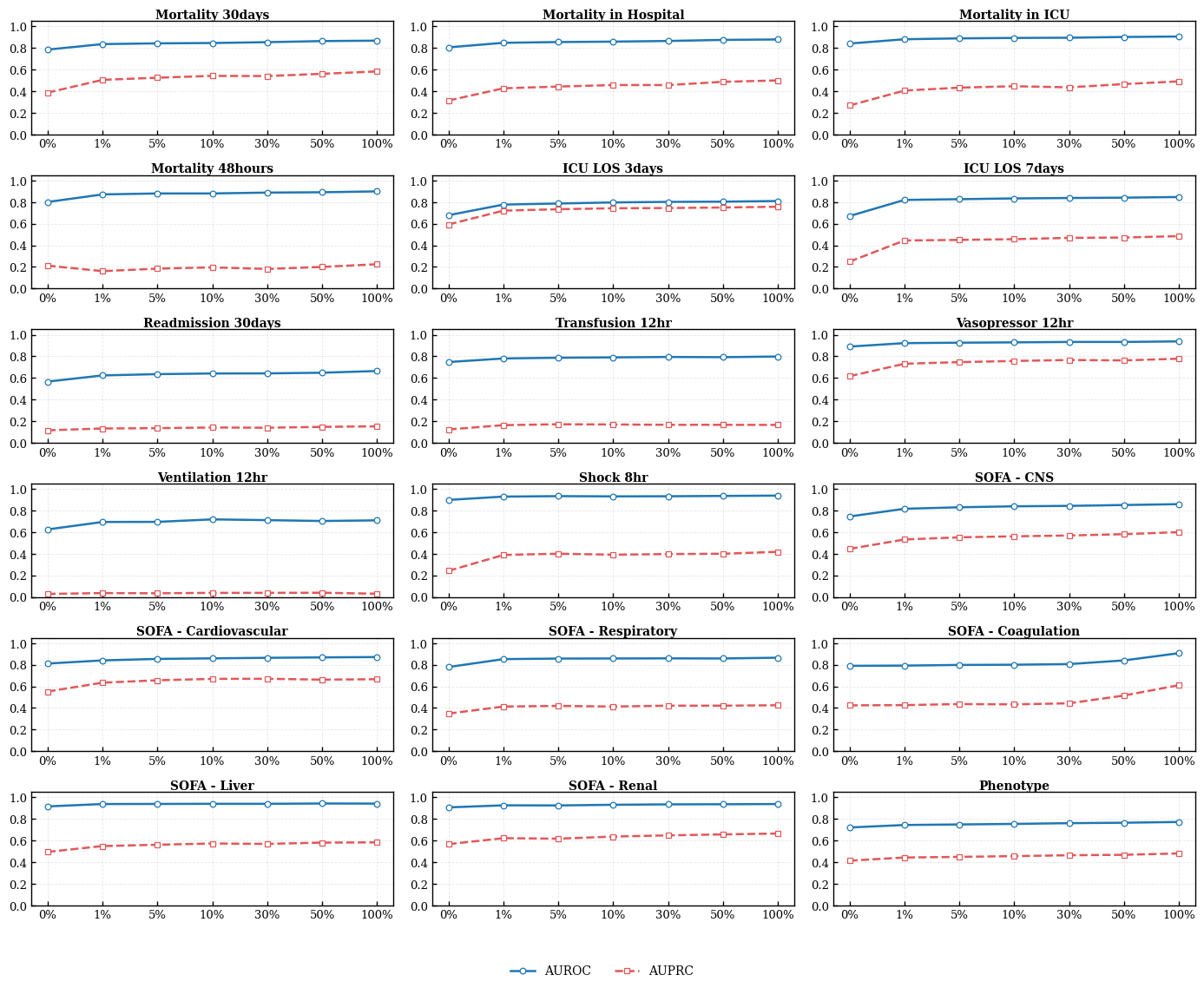}
    \caption{
Performance comparison of the limited-variable version of the model on the MIMIC-IV YAIB datasets across varying fine-tuning data ratios.
Each subplot illustrates the AUROC and AUPRC trends for binary, multi-class (SOFA), and multi-label (Phenotype) clinical prediction tasks when using a reduced set of variables common to both datasets.
}
    \label{figure : MIMIC YAIB ver All Tasks}
\end{figure*}

\FloatBarrier

\begin{figure*}[ht]
    \centering
    \includegraphics[width=0.9\textwidth]{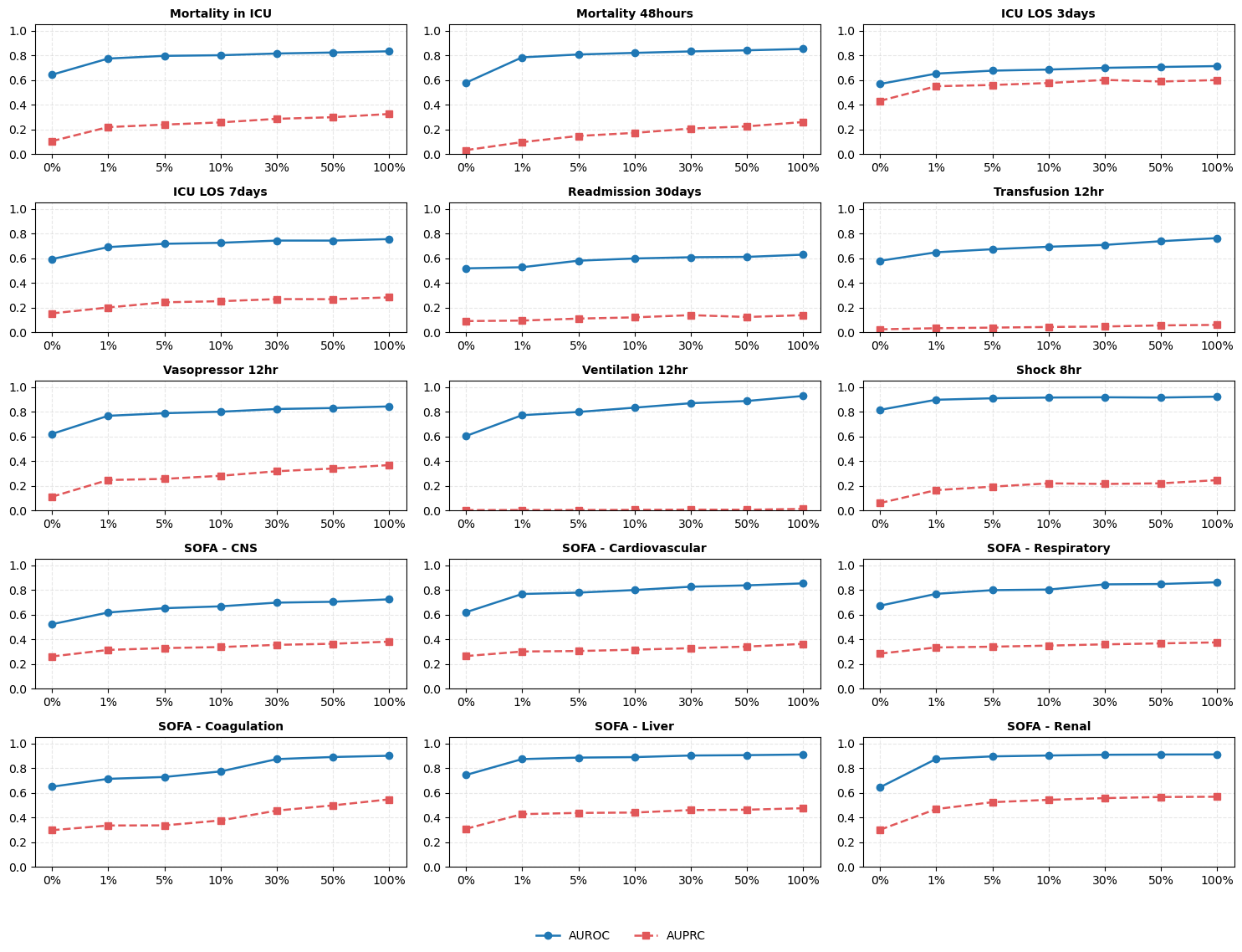}
    \caption{
External validation performance of the model on the eICU dataset.
AUROC and AUPRC are reported for 9 binary clinical prediction tasks and 6 SOFA multi-class tasks under varying fine-tuning data ratios (0\%–100\%).
}
    \label{figure : eICU external validation}
\end{figure*}

\FloatBarrier

\begin{figure*}[ht]
    \centering
    \includegraphics[width=0.9\textwidth]{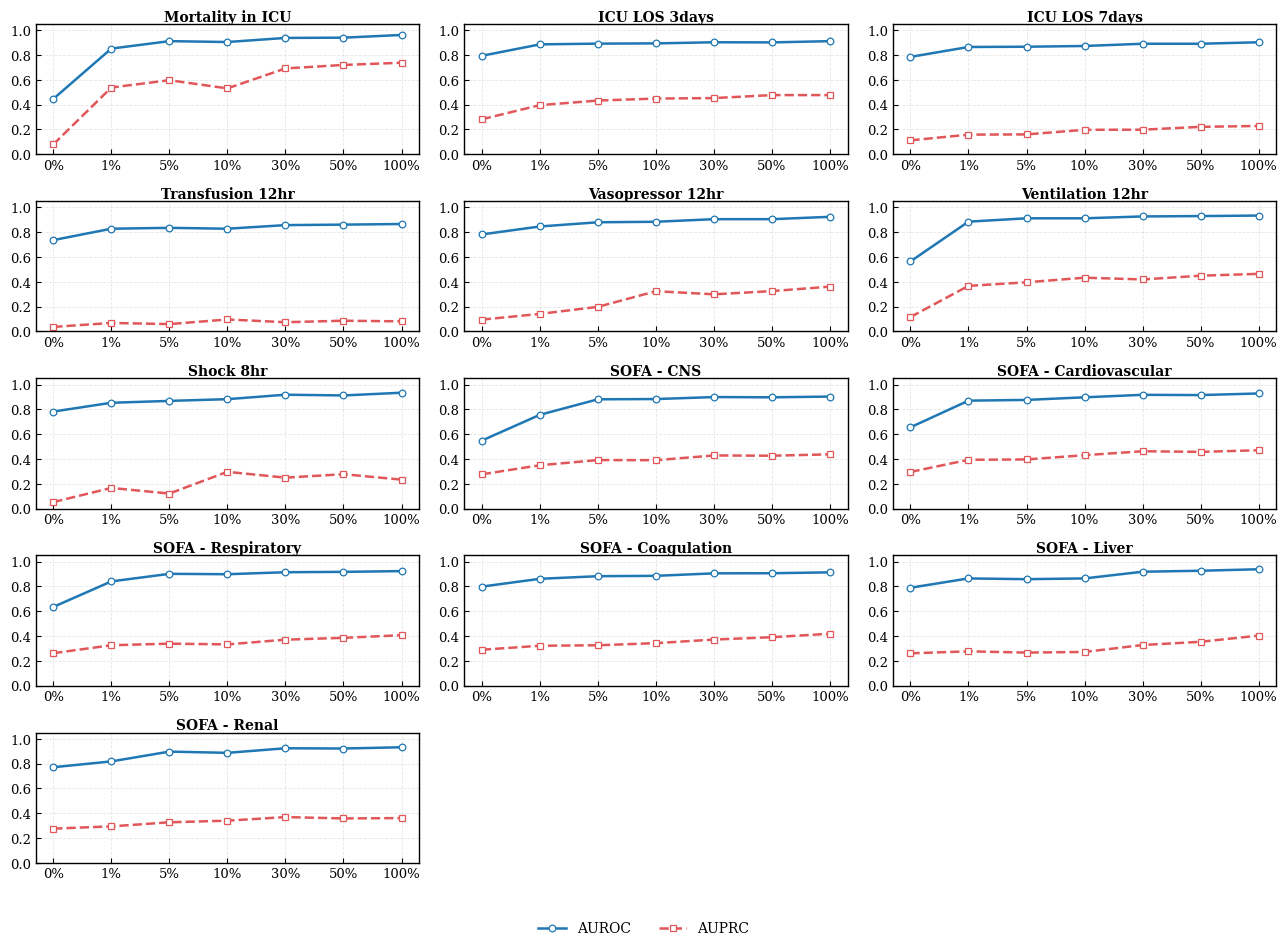}
    \caption{
External validation performance of the model on the HiRID dataset (first version).
AUROC and AUPRC are reported for 9 binary clinical prediction tasks and 6 SOFA multi-class tasks under varying fine-tuning data ratios (0\%–100\%).
}
    \label{figure : HiRID external validation first}
\end{figure*}

\FloatBarrier

\begin{figure*}[ht]
    \centering
    \includegraphics[width=0.9\textwidth]{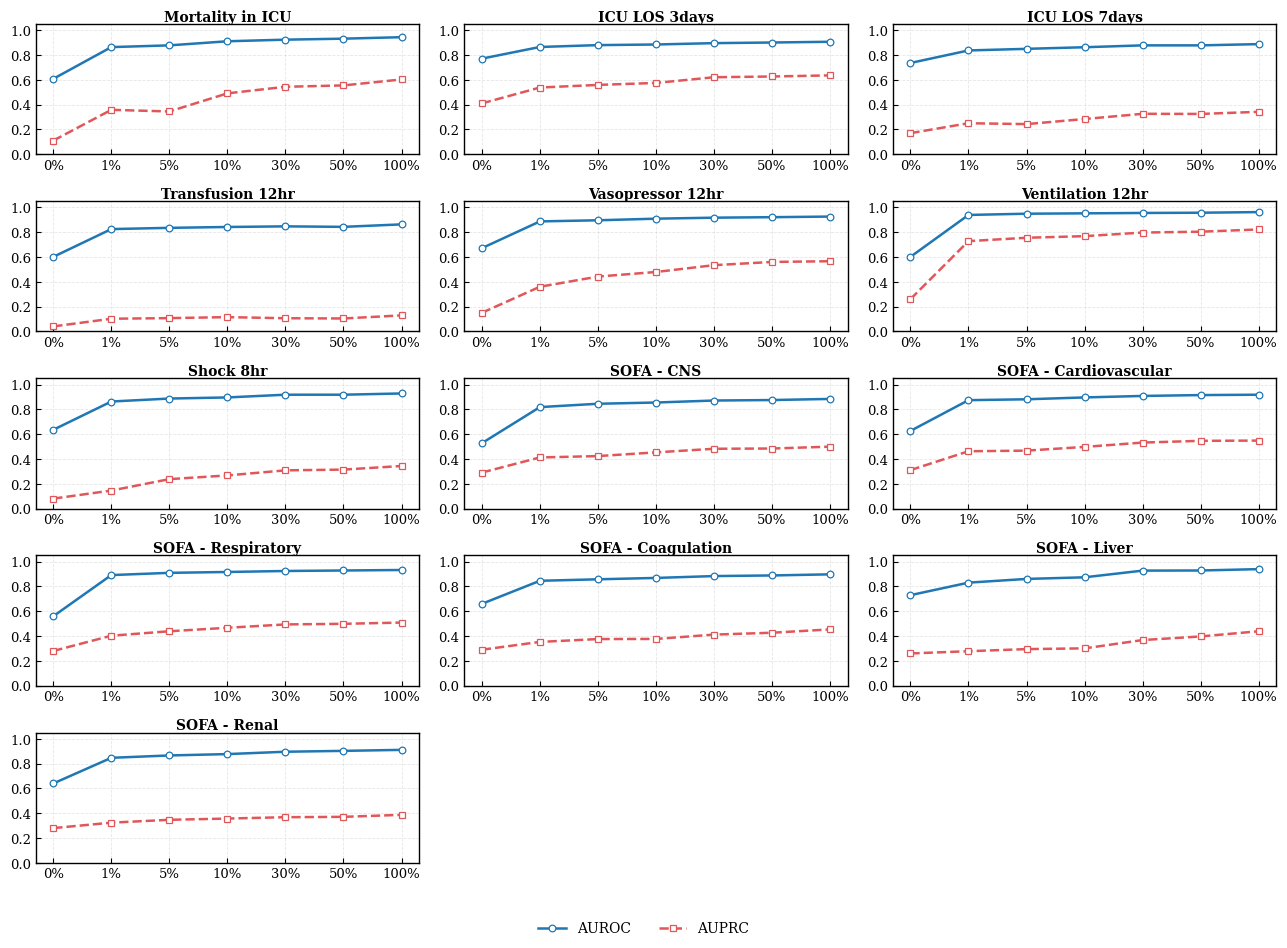}
    \caption{
External validation performance of the model on the HiRID dataset (last version).
AUROC and AUPRC are reported for 9 binary clinical prediction tasks and 6 SOFA multi-class tasks under varying fine-tuning data ratios (0\%–100\%).
}
    \label{figure : HiRID external validation last}
\end{figure*}

\FloatBarrier

\begin{figure*}[ht]
    \centering
    \includegraphics[width=0.9\textwidth]{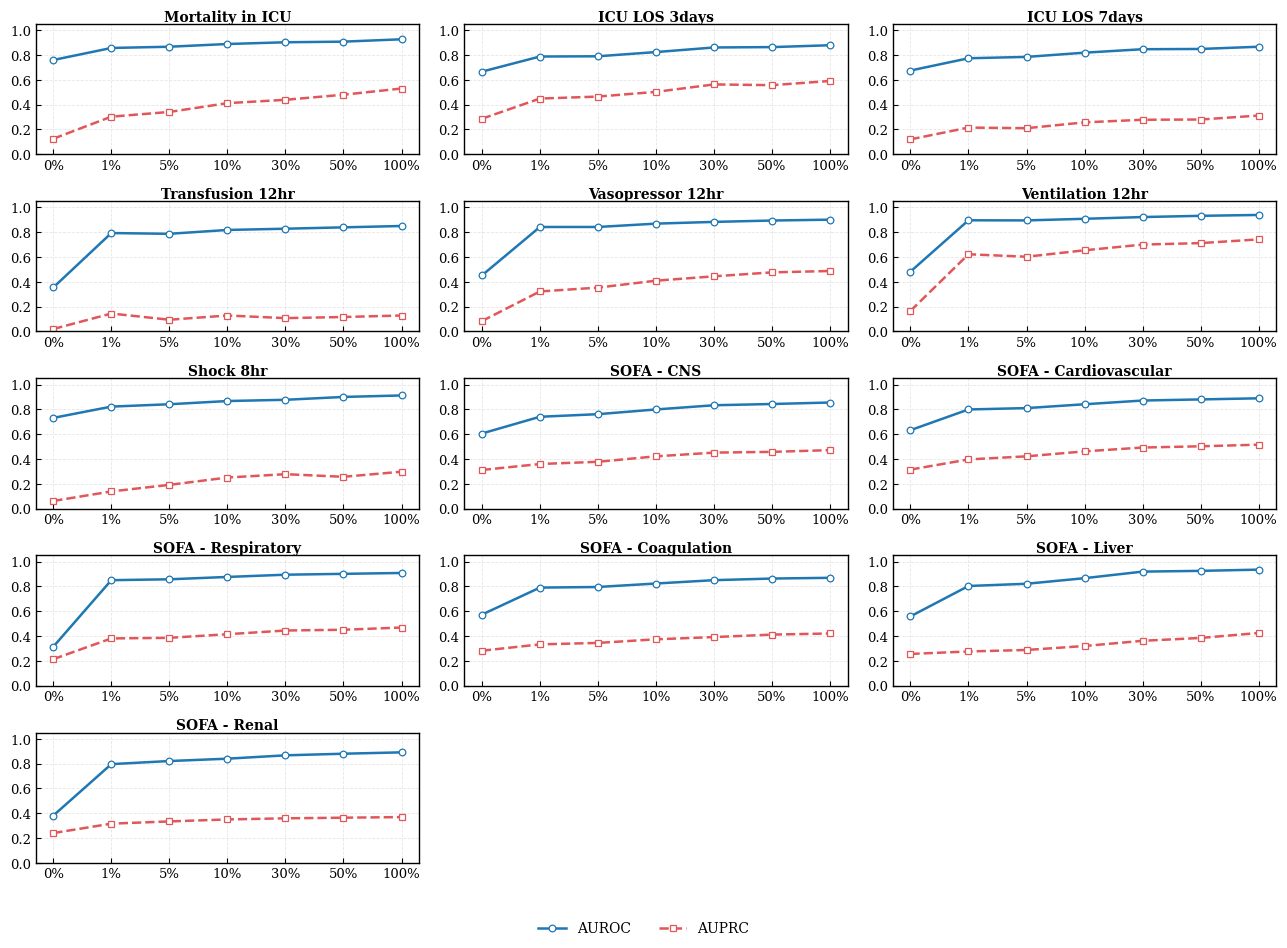}
    \caption{
External validation performance of the model on the HiRID dataset (4093-token version).
AUROC and AUPRC are reported for 9 binary clinical prediction tasks and 6 SOFA multi-class tasks under varying fine-tuning data ratios (0\%–100\%).
}
    \label{figure : HiRID external validation 4093}
\end{figure*}

\FloatBarrier

\begin{figure*}[ht]
    \centering
    \includegraphics[width=0.9\textwidth]{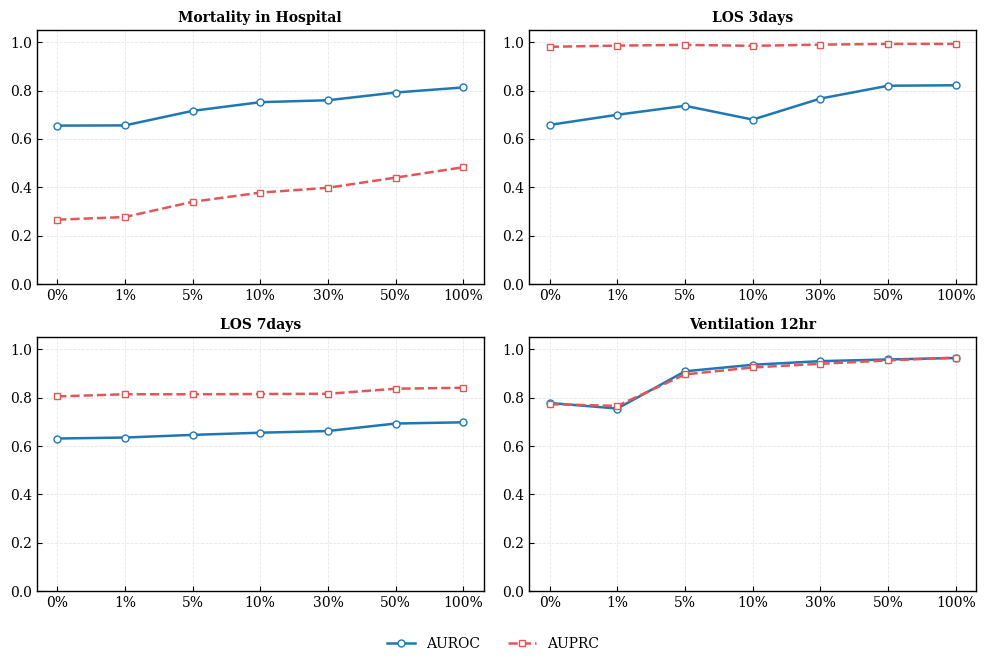}
    \caption{
Performance of the model on the PhysioNet 2012 (P12) dataset with a 24-hour observation window across varying fine-tuning data ratios.
Each subplot presents AUROC and AUPRC results for representative clinical prediction tasks — including Mortality in Hospital, Length-of-Stay (3/7 days), and Ventilation within 12 hours.
}
    \label{figure : P12 external validation 24 window}
\end{figure*}

\FloatBarrier

\begin{figure*}[ht]
    \centering
    \includegraphics[width=0.9\textwidth]{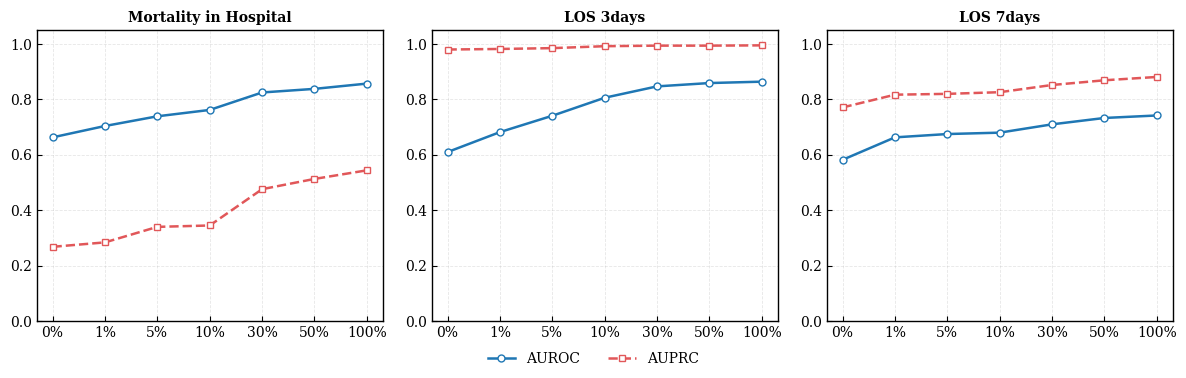}
    \caption{
Performance of the model on the PhysioNet 2012 (P12) dataset with a 48-hour observation window.
AUROC and AUPRC are presented for three clinical prediction tasks — Mortality in Hospital, Length of Stay (3/7 days) — across increasing fine-tuning data ratios.
The Ventilation 12hr task was excluded in this window. 
}
    \label{figure : P12 external validation 48 window}
\end{figure*}

\FloatBarrier

\section*{Data Availability}

All datasets used in this study are publicly available under appropriate data use agreements.

The \textbf{MIMIC-IV} (version~2.2) database~\cite{johnson2023mimic} is available through the PhysioNet repository 
(\url{https://physionet.org/content/mimiciv/2.2/}) 
upon completion of the required credentialing process and data use agreement.

The \textbf{HiRID} (version~1.1.1) dataset~\cite{hyland2020early} is publicly accessible from PhysioNet 
(\url{https://physionet.org/content/hirid/1.1.1/}) 
and may be used for research under a corresponding data use license.

The \textbf{eICU Collaborative Research Database}~\cite{pollard2018eicu} 
can be accessed through PhysioNet 
(\url{https://physionet.org/content/eicu-crd/2.0/}) 
after completion of the credentialed user training and signing of the data use agreement.

The \textbf{PhysioNet Challenge 2012 (P12)} dataset~\cite{silva2012predicting} is also available via PhysioNet 
(\url{https://physionet.org/content/challenge-2012/1.0.0/}) 
and may be freely used for research and reproducibility purposes.

\section*{Code Availability}
The underlying code for this study is available at: https://github.com/sejeongak/PULSE-ICU

\bibliography{sn-bibliography}

\section*{Acknowledgements}
This research work was funded in part by the National Research Foundation (NRF) grant funded by the Korea government (MSIT) (No. RS-2023-00212713) (received by Lee HK). This study was also partially supported by following grants: National Institutes of Health R35GM159939, Korea Health Industry Development Institute RS-2024-00439677 (received by Yoon JH). This work was supported by
the MSIT (Ministry of Science and ICT), Korea, under the ICAN (ICT Challenge and Advanced Network of HRD) support program (IITP-2024-00438411) supervised by the IITP (Institute for Information \& Communications Technology Planning \& Evaluation). This work was supported in part by the BK21 FOUR funded by the Ministry of Education of Korea and National Research Foundation of Korea.

\section*{Author contributions}
Conceptualization and methodology: SJ, JHY, and HKL. Data analysis and model development: SJ. Supervision: JHY and HKL. Writing original draft and revision: SJ, JHY, and HKL.

\section*{Competing interests}
The authors declare no competing interests.

\section*{Additional information}
Not applicable.

\end{document}